\documentclass{article}

\usepackage{iclr2027_conference,times} 

\usepackage{amsmath,amsfonts,bm}

\def\eqref#1{equation~\ref{#1}}

\def\1{\bm{1}}

\DeclareMathAlphabet{\mathsfit}{\encodingdefault}{\sfdefault}{m}{sl}
\SetMathAlphabet{\mathsfit}{bold}{\encodingdefault}{\sfdefault}{bx}{n}

\usepackage{times}
\usepackage{latexsym}
\usepackage[T1]{fontenc}
\usepackage[utf8]{inputenc}

\usepackage{amsmath}
\usepackage{amssymb}
\usepackage{booktabs}
\usepackage{graphicx}
\usepackage{wrapfig}
\usepackage{xcolor}
\usepackage{colortbl}
\usepackage{multirow}
\usepackage{hyperref}
\hypersetup{hidelinks}
\usepackage{microtype}
\usepackage{url}
\usepackage{enumitem}

\usepackage[most]{tcolorbox}
\usepackage{xspace}

\def\method{{\sc T-Mem}\xspace}

\definecolor{caseheader}{RGB}{82,82,82}
\definecolor{caseblue}{RGB}{43,91,178}
\definecolor{casepink}{RGB}{184,62,115}
\definecolor{casegreen}{RGB}{42,120,83}
\definecolor{caselight}{RGB}{247,248,250}

\newtcolorbox{casebox}[1]{
    enhanced,
    width=\linewidth,
    colback=white,
    colframe=caseheader,
    colbacktitle=caseheader,
    coltitle=white,
    title={#1},
    fonttitle=\bfseries\small,
    boxrule=0.7pt,
    arc=1.5mm,
    outer arc=1.5mm,
    left=2.6mm,
    right=2.6mm,
    top=2.0mm,
    bottom=2.0mm,
    before skip=3pt,
    after skip=2pt
}

\newtcolorbox{evidencebox}{
    enhanced,
    colback=caselight,
    colframe=black!20,
    boxrule=0.4pt,
    arc=0.8mm,
    left=1.8mm,
    right=1.8mm,
    top=1.3mm,
    bottom=1.3mm
}

\title{Temporal Evidence Forests for Long-Horizon Multi-Party Conversational Memory}

\author{
Junyou Guo\textsuperscript{1,\ddag} \quad Chongyang Tao\textsuperscript{1,\dag} \quad Yan He\textsuperscript{1} \quad Yong Wang\textsuperscript{2} \quad Shuai Ma\textsuperscript{1}\\[4pt]
{\footnotesize \textsuperscript{1}School of Computer Science and Engineering, Beihang University, Beijing, China}\\[2pt]
{\footnotesize \textsuperscript{2}Alibaba Group}\\[2pt]
{\footnotesize \textsuperscript{\ddag}Project Lead. \quad \textsuperscript{\dag}Corresponding author.}
}

\iclrfinalcopy

\makeatletter
\renewcommand{\@maketitle}{%
  \vbox{\hsize\textwidth
    \centering
    {\LARGE\scshape \@title\par}
    \vskip 16pt
    {\normalsize\normalfont
      \begin{tabular}[t]{c}\@author\end{tabular}\par}
    \vskip 24pt
  }%
}
\makeatother
\renewcommand{\headrulewidth}{0pt}

\begin{document}

\maketitle

\begin{abstract}
Long-term memory has been widely adopted in LLM-based agents to support
persistent interaction and personalized assistance. Long-term multi-party
interactions, however, accumulate substantially more detailed information,
with decisions repeatedly updated and states evolving across participants and
interaction stages. Temporal structure is therefore crucial for tracing such
memories: relevant information often appears densely within short intervals,
while transitions among different states are anchored by a strict temporal
order. Existing temporally aware memory frameworks primarily store time as
metadata or use temporal signals to guide retrieval over preconstructed memory
structures, without using temporal structure to both recover locally concentrated
evidence and reconstruct its complete event progression. To address
these limitations, we propose \method, a temporal-structure-based memory framework
that preserves interactions and organizes query-relevant evidence through two
temporal mechanisms. Temporal Evidence Discovery exploits the local
concentration of related messages to recover overlooked evidence, while
Temporal Memory Structuring models state transitions and supporting relations
over time and compresses low-information spans within the resulting forest.
Evidence Forest Linearization then serializes the structured evidence for
answer generation. Experiments on two long-term multi-party memory benchmarks
demonstrate that \method consistently outperforms competitive memory
frameworks across LLMs, parameter settings, domains, and query types.
\end{abstract}

\section{Introduction}
\label{sec:introduction}

An LLM agent operating over a long interaction history cannot keep every prior message in its active context. External-memory systems therefore preserve past interactions as passages, summaries, or structured records and retrieve a compact subset before generation \citep{lewis2020rag,packer2023memgpt,xu2026memgas,latimer2026hindsight,xu2026structmem}. In long-term multi-party interactions, however, the challenge is not only scale. Parallel discussions are interleaved across speakers and stages, while the same issue may be repeatedly proposed, revised, rejected, or confirmed. Semantically similar messages can therefore describe incompatible states, whereas a decisive update may share few words with the eventual query. A useful memory context must identify the temporally coherent episode and expose the complete state evolution that determines the current answer.

We identify two complementary temporal properties that remain poorly represented by existing memory frameworks. The first is \emph{temporal locality}. Memories contributing to the same event often concentrate within a discussion episode, even when an individual message has little surface overlap with a later query. A retrieved message can thus serve not only as evidence but also as a temporal anchor for locating its surrounding episode. Simply increasing Top-$k$ may recover additional evidence, but it also introduces unrelated episodes and competing topics. The question is therefore not only which messages are relevant, but also where the relevant interaction is temporally concentrated.

The second property is \emph{temporal evolution}, which determines how recovered evidence should be interpreted and presented. Within an episode, messages are not interchangeable observations: a proposal may be questioned, refined, approved, implemented, or replaced, while another message may explain or support a later decision. A similarity-ranked list can retrieve several messages about the same subject without revealing which one represents the current state or how they form an event trajectory. Correct answering therefore requires reconstructing the progression connecting these messages rather than treating them as independent evidence items.

Evidence completeness alone is therefore insufficient. When provisional proposals, later revisions, and supporting explanations are presented as an unordered set, the answer model must reconstruct both their chronology and their functional roles from scratch. Earlier and superseded states may then compete with the final decision despite all required messages being present. Explicitly distinguishing state-changing updates from supporting elaborations instead converts isolated matches into an interpretable trajectory and makes the evolution leading to the answer directly observable.

Existing agent-memory systems improve access through graph-based organization, persistent memory networks, multi-granularity association, and selective consolidation \citep{gutierrez2025hipporag2,rasmussen2025zep,chhikara2025mem0,jiang2026magma}. These approaches are generally centered on semantic association, entity structure, or pre-query consolidation. Although temporal information may be retained, it does not jointly serve as the primary basis for locating a relevant episode and reconstructing its internal state transitions. Consequently, even a relevant memory set may be presented as disconnected records or transformed facts, leaving the answer model to infer which messages belong to the same event and which state supersedes another. This limitation is particularly consequential in long-term group conversations, where sophisticated memory systems do not consistently outperform simple retrieval \citep{yang2026groupmembench}. Temporal locality and temporal evolution address complementary failure modes: the former determines where coherent evidence is concentrated, while the latter determines how the recovered messages should be interpreted. Both are needed to recover an interaction episode and expose its complete evolution for generation.

We introduce \textbf{\method}, a temporal-structure-aware memory framework for long-term multi-party interactions, built around these two objectives. \emph{Temporal Evidence Discovery} exploits temporal locality: Seed Retrieval first obtains anchors, whose temporal distribution identifies concentrated, topic-specific interaction episodes. \method then searches the original messages within these episodes to recover answer-critical evidence that global similarity ranking may overlook. \emph{Temporal Memory Structuring} models temporal evolution by organizing the recovered messages into a query-conditioned evidence forest with chronologically constrained \textsc{Update} and \textsc{Expand} relations. \textsc{Update} captures state progression, whereas \textsc{Expand} captures answers, explanations, and supporting evidence. After these relations are established, low-information spans are compressed, and Evidence Forest Linearization presents the ordered messages and their relations to the answer model. The resulting context exposes the complete event progression rather than a flat collection of memories.

We evaluate \method on GroupMemBench \citep{yang2026groupmembench} and EverMemBench \citep{hu2026evermembench}. \method achieves 63.09\% and 62.68\% overall accuracy on GroupMemBench with GPT-5.5 and Deepseek-V4-Flash, and 79.46\% and 77.63\% on EverMemBench, respectively. Across both benchmarks, \method consistently outperforms flat retrieval and strong agent-memory baselines.
Controlled comparisons support the importance of recovering temporally coherent episodes and explicitly presenting their state evolution. 
Our contributions are fourfold:
\begin{itemize}[itemsep=2pt, topsep=2pt, parsep=3pt, partopsep=5pt]
  \item We introduce \method, an agent-memory framework that treats temporal structure as a core representation for retrieving and organizing evidence in long-term multi-party interactions.

  \item We propose Temporal Evidence Discovery, which exploits temporal locality to locate concentrated interaction episodes around retrieved anchors and recover answer-critical messages overlooked by global similarity ranking.

  \item We develop Temporal Memory Structuring and Evidence Forest Linearization, which reconstruct chronologically grounded relations and present the complete query-conditioned event evolution to the answer model.

  \item We conduct extensive experiments on GroupMemBench and EverMemBench, where \method achieves the highest overall accuracy across two LLM backbones and outperforms the strongest evaluated baseline by 6.50--8.59 percentage points.
\end{itemize}

\vspace{-2mm}
\section{Related Work}
\label{sec:related_work}

\vspace{-1.5mm}
\paragraph{Structured and Retrieval-Augmented Memories}
Retrieval-augmented memory systems select information from external stores to supplement the limited context of language models. Lexical and dense retrieval provide efficient access to relevant records, while structured approaches expose broader associations and support multi-step reasoning. RAPTOR~\citep{sarthi2024raptor} recursively summarizes documents into a retrieval tree, whereas GraphRAG~\citep{edge2024graphrag} constructs entity graphs and community summaries for query-focused generation. HippoRAG~\citep{gutierrez2024hipporag} and HippoRAG2~\citep{gutierrez2025hipporag2} combine knowledge graphs with associative retrieval to connect related facts. Agent-memory systems further transform interaction histories into reusable units: MemGPT~\citep{packer2023memgpt} manages information through hierarchical memory tiers and context swapping, while MemGAS~\citep{xu2026memgas} constructs multi-granularity associations and adaptively selects the retrieval granularity. These approaches broaden the information and relations available at inference time, but mainly determine which memory units are relevant. Long-term conversations also require determining when information applies and how it changes across interactions, motivating temporally aware memory.

\vspace{-1.5mm}
\paragraph{Temporally Aware Memories}
Temporally aware memory systems incorporate temporal cues across the memory lifecycle. A-Mem~\citep{xu2025amem} incrementally evolves interconnected notes as new experiences arrive, allowing its memory representation to adapt over time. Zep~\citep{rasmussen2025zep} represents facts and their validity intervals in a temporal knowledge graph for time-aware retrieval. Hindsight~\citep{latimer2026hindsight} organizes memories through temporal and entity-aware networks and combines semantic, lexical, graph-based, and temporal retrieval signals. StructMem~\citep{xu2026structmem} temporally anchors event-level memories before periodically consolidating semantically related events into cross-event representations. PRISM~\citep{peng2026prism} performs intent-aware retrieval and compression over a preconstructed memory graph, routing different query intents, including temporal queries, through typed relation paths. Although these systems incorporate time at different stages, it generally serves as an auxiliary memory attribute or a signal for navigating structures organized by other relations. Consequently, they do not explicitly organize retrieved memories into the complete progression of an event. Furthermore, they do not exploit temporal locality to recover answer-critical evidence concentrated around relevant interaction episodes.

\vspace{-1.5mm}
\paragraph{Memory in Long-Term Multi-Party Conversations}
GroupMemBench~\citep{yang2026groupmembench} and EverMemBench~\citep{hu2026evermembench} extend long-term memory evaluation to multi-party and collaborative conversations. Unlike conventional long-term dialogues, these conversations interleave multiple speakers, roles, topics, and concurrently evolving decisions. Relevant evidence may therefore be distributed across participants and interaction stages, while lexically similar messages can describe different episodes or conflicting states. Addressing this setting requires more than retrieving individually relevant memories: the system must identify when the relevant interaction occurred and organize its messages according to their temporal and evidential relationships. \method addresses these challenges by discovering temporally localized evidence and structuring the original messages into query-conditioned evidence forests, providing the answer model with a coherent representation of multi-party state evolution.

\section{Problem Formulation}
\label{sec:problem}

Consider a group-chat memory $\mathcal{M}=\{m_i\}_{i=1}^{N}$. Each message is
a tuple $m_i=(x_i,t_i,z_i)$ containing text $x_i$, timestamp $t_i$, and
conversation metadata $z_i$, such as author, role, channel, topic, and phase.
A query $q=(x_q,u)$ consists of query text $x_q$ and the requesting user $u$.
For each query, the task is to derive a compact evidence context
$\mathcal{S}_q$ from $\mathcal{M}$ and generate an answer
$\hat{y}_q=g(q,\mathcal{S}_q)$ that is supported by the conversation.
Reference answers and evidence annotations are used only for evaluation and are
unavailable at inference time.

Let $L(\mathcal{S}_q)$ denote the number of model-input tokens in the evidence
context, $B$ is the maximum permitted token budget. Given an evaluation
set $\mathcal{Q}$ with reference answers $\{y_q\}_{q\in\mathcal{Q}}$, the
objective is
\begin{equation}
  \max_{\{\mathcal{S}_q\},g}
  \mathbb{E}_{q\sim\mathcal{Q}}\!\left[
    J\!\left(g(q,\mathcal{S}_q),y_q\right)
  \right]
  \quad \text{s.t.} \quad
  L\!\left(\mathcal{S}_q\right)\leq B,\;\forall q\in\mathcal{Q},
  \label{eq:task_objective}
\end{equation}
where $J(\hat{y}_q,y_q)\in\{0,1\}$ denotes the benchmark correctness
judgment. This formulation defines long-term multi-party memory question
answering under a bounded input context without prescribing how the evidence
context is obtained.

\section{\method}
\label{sec:method}

\subsection{Overview}

\begin{figure*}[t]
  \centering
  \vspace{-4mm}
  \includegraphics[width=\textwidth]{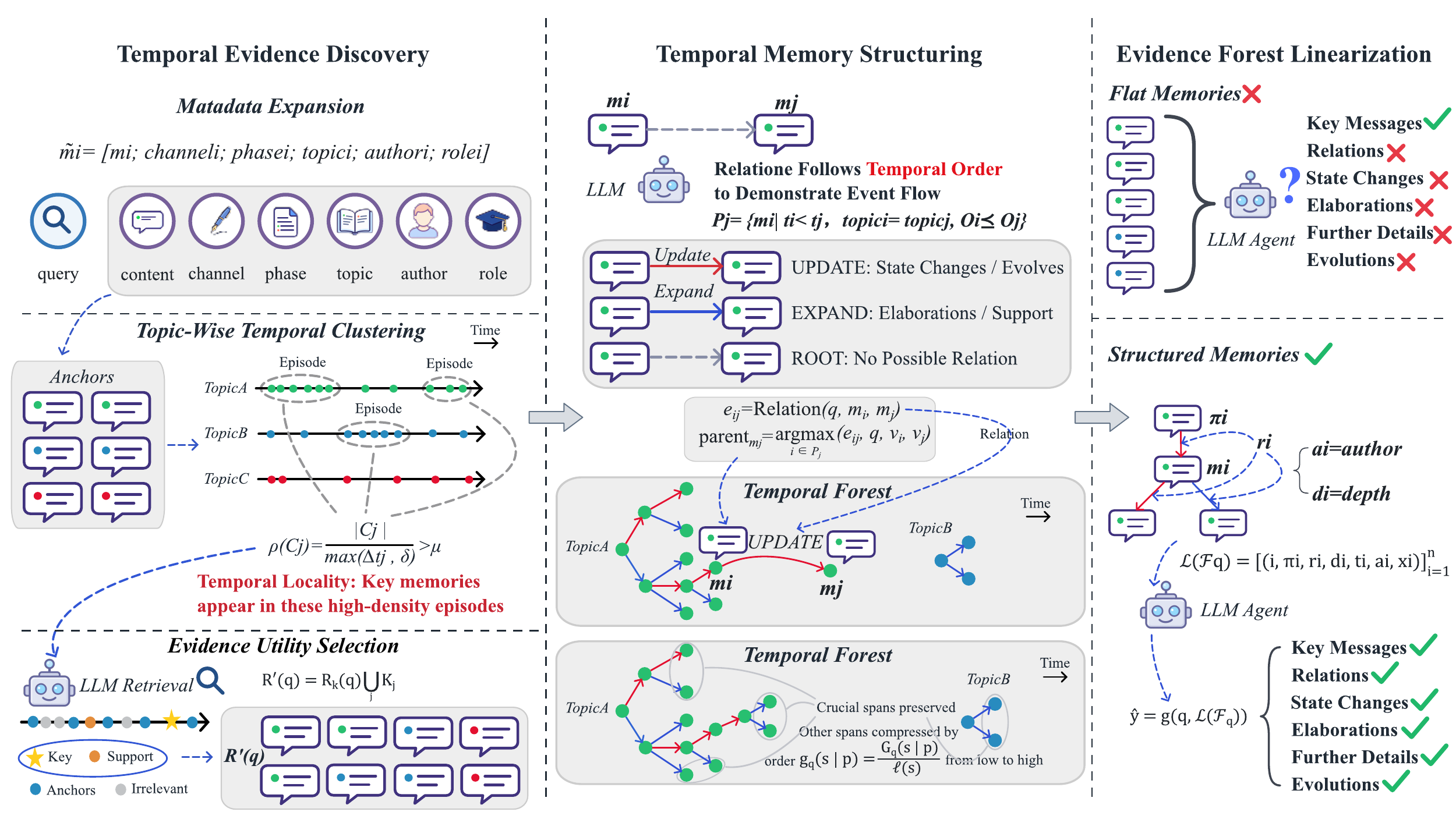}
  \vspace{-5mm}
  \caption{\textbf{Overview of \method.} Metadata-expanded BM25 and Temporal Evidence Discovery recover candidate evidence; Temporal Memory Structuring links state evolution and support while compressing low-information spans; and Evidence Forest Linearization presents the resulting forest.}
  \vspace{-3mm}
  \label{fig:architecture}
\end{figure*}

\method leaves the stored conversation unchanged and builds a compact
evidence view when a query arrives. The three stages in
Figure~\ref{fig:architecture} discover candidate evidence, structure state changes and support, and linearize the resulting forest. For memory
$\mathcal{M}=\{m_i\}_{i=1}^{N}$ and query $q$, the output is a forest
$\mathcal{F}_q=(\mathcal{V}_q,\mathcal{E}_q)$ whose nodes are original messages.
The answer model receives $q$ and a serialization of this forest. It never sees
the reference answer.
Metadata-expanded Seed Retrieval first supplies coarse anchors, and Temporal Evidence
Discovery searches the corresponding episodes to recover memories that exhibit
temporal locality with these anchors. Temporal Memory Structuring then links
state-changing and supporting relations under strict chronological constraints
and compresses low-information spans within the resulting forest. Evidence
Forest Linearization converts this structure into the answer context.

\subsection{Temporal Evidence Discovery}
\label{sec:temporal_expansion}

Group-chat events often unfold within temporally concentrated episodes containing
proposals, confirmations, and subsequent updates. We first retrieve query-aligned
seed messages using a lightweight retriever (e.g., BM25), representing each
message by its text together with available channel, phase, topic, author, and
role metadata. The top-$k$ messages form the coarse anchor set $R_k(q)$. Beyond
serving as direct evidence, these seeds provide temporal anchors for locating
answer-critical messages that may have low global relevance. We therefore use
their temporal distribution to identify candidate evidence intervals within each
topic and inspect the original conversations in those intervals.

\vspace{-2mm}
\paragraph{Topic-wise temporal clustering.}
For each topic $\theta$ represented in $R_k(q)$, we define the topic-specific seed set as$
A_\theta=\{m_i\in R_k(q):\mathrm{topic}_i=\theta\}.
\label{eq:topic_seed_set}
$.
We group the messages in $A_\theta$ into temporal clusters according to the proximity of their timestamps (e.g., DBSCAN or agglomerative clustering). Cluster boundaries are determined by the temporal clustering procedure adopted in implementation. For a resulting cluster $C_j\subseteq A_\theta$ with temporal span $\Delta t_j$, we measure its seed density as
\begin{equation}
  \rho(C_j)=\frac{\lvert C_j\rvert}
  {\max(\Delta t_j,\delta)},
  \label{eq:density}
\end{equation}
where $\delta>0$ prevents an extremely short interval from receiving an unbounded score. Clusters satisfying $\rho(C_j)>\mu$ define candidate evidence intervals, which are padded to include nearby confirmations, corrections, and follow-up updates. The numerator favors intervals containing multiple query-relevant seeds, while the temporal span penalizes diffuse activity.

\vspace{-1.5mm}
\paragraph{Evidence Utility Selection.}
Content similarity provides a first filter within each window. A
query-conditioned LLM then labels the remaining messages as \textsc{Key},
\textsc{Support}, or \textsc{Irrelevant}. \textsc{Key} messages establish answer
claims; \textsc{Support} messages provide necessary bridges or constraints. The
accepted set $K_j$ contains both labels from window $j$:
\begin{equation}
  R'(q)=R_k(q)\cup\bigcup_j K_j.
  \label{eq:expanded_retrieval}
\end{equation}
Similarity is used for recall, but acceptance depends on evidence utility for
the query. The gate therefore does not copy a full discussion merely because its
messages share keywords or a topic.

\subsection{Temporal Memory Structuring}
\label{sec:forest}
\label{sec:compression}

\method orders $R'(q)$ by time and builds a temporary, query-conditioned forest over the
original messages. A possible parent of $m_j$ must be earlier, belong to the same
topic, and respect the conversation-stage order:
\begin{equation}
  \begin{aligned}
  \mathcal{P}_j&=\{m_i\mid t_i<t_j,\ \text{topic}_i=\text{topic}_j,\
  O_i\preceq O_j\},\\
  O_i&=(\text{stage}_i,\text{phaseOrder}_i,\text{phaseProgress}_i).
  \end{aligned}
  \label{eq:parent_candidates}
\end{equation}
The tuple $O_i$ collects the ordered stage and phase fields in $z_i$;
$\preceq$ is lexicographic order. A query-conditioned classifier assigns each
eligible pair one of three labels:
\begin{equation}
e_{ij}=\operatorname{Relation}(q,m_i,m_j)=\begin{cases}
\textsc{Update}, & \text{if $m_j$ advances or changes the state in $m_i$},\\
\textsc{Expand}, & \text{if $m_j$ answers, explains, or directly supports $m_i$},\\
\textsc{Root}, & \text{if neither relation is supported.}
\end{cases}
\label{eq:relations}
\end{equation}
The highest-confidence non-\textsc{Root} candidate becomes the parent; otherwise,
$m_j$ starts a new tree. Temporal, topic, and stage constraints prevent backward,
cross-topic, or progression-reversing links. An \textsc{Update} edge records state
progression or replacement, whereas an \textsc{Expand} edge adds support without
superseding its parent.

The same module compresses node context when the serialized forest exceeds the
answer budget $B$, without deleting nodes or changing edges. Each leaf $v$ is
paired with its parent $p$ and divided into spans $\mathcal{S}_{p,v}$. Let $U(s)$
denote the normalized lexical units in span $s$, $N$ the number of corpus
messages, and $\operatorname{df}(u)$ the number containing unit $u$. We define
\begin{equation}
  G_q(s\mid p)
  =
  \sum_{u\in U(s)\cap U(q)}
  \left(\log\frac{N+1}{\operatorname{df}(u)+1}+1\right)
  +
  \sum_{u\in U(s)\setminus U(p)}
  \left(\log\frac{N+1}{\operatorname{df}(u)+1}+1\right),
  \label{eq:information_gain}
\end{equation}
which rewards query-relevant content and information not already conveyed by
the parent. Its token-normalized score is
\begin{equation}
  g_q(s\mid p)=\frac{G_q(s\mid p)}{\ell(s)},
  \label{eq:information_density}
\end{equation}
where $\ell(s)$ denotes the span length. Spans containing query entities, timestamps, numerical values, or state changes along \textsc{Update} edges are protected. The remaining spans are omitted in ascending order of $g_q(s\mid p)$ until the retained content satisfies $\sum_{s\in Z}\ell(s)\leq B$. If necessary, compression proceeds iteratively from leaves to newly exposed parent nodes. Every node retains its metadata, relation label, and at least one text span.


\subsection{Evidence Forest Linearization}
\label{sec:forest-linearization}

We convert the structured evidence forest into a sequence that can be directly processed by the answer model. Trees are ordered by their root timestamps, while nodes within each tree follow a preorder traversal with siblings arranged chronologically.
For each node $v_i$, let $\pi_i$ denote the index of its parent and $r_i$ the corresponding relation. A root has $\pi_i=0$ and $r_i=\textsc{Root}$; otherwise, $r_i\in\{\textsc{Update},\textsc{Expand}\}$. The linearized forest is
\begin{equation}
\mathcal{L}(\mathcal{F}_q)
=
\left[
(i,\pi_i,r_i,d_i,t_i,a_i,x_i)
\right]_{i=1}^{n},
\end{equation}
where $d_i$, $t_i$, $a_i$, and $x_i$ denote the depth, timestamp, author, and retained text of $v_i$. Encoding each node together with its parent and relation preserves the forest structure and temporal order in a model-readable sequence. The answer model then generates the final response as 
$
\hat{y}
=
g
\left(q,\mathcal{L}(\mathcal{F}_q)\right).
$

\section{Experiments}
\label{sec:experiments}

\subsection{Experimental Setup}
\label{sec:experimental_setup}

\paragraph{Datasets and Metrics.}
We evaluate on GroupMemBench~\citep{yang2026groupmembench} and
EverMemBench~\citep{hu2026evermembench}, following their official data
and evaluation protocols. Both benchmarks test long-term memory in multi-party
conversations. We report answer accuracy as the primary metric.

\paragraph{Baselines.}
We compare \method with a diverse set of strong baselines, including BM25, Qwen3.7-text-embedding~\citep{zhang2025qwen3}, A-Mem~\citep{xu2025amem}, HippoRAG2~\citep{gutierrez2025hipporag2}, Hindsight~\citep{latimer2026hindsight}, StructMem~\citep{xu2026structmem}, and MemGAS~\citep{xu2026memgas}. These baselines span conventional lexical and dense retrieval, retrieval-augmented memory, and recent structured agent-memory systems, providing broad coverage of representative approaches to long-term memory. Detailed baseline configurations are provided in Appendices~\ref{app:baselines} and~\ref{app:settings}.


\paragraph{Implementation Details.}
For Temporal Evidence Discovery, we use BM25 to retrieve
$k=36$ coarse anchors. We instantiate the temporal clustering module with DBSCAN, clustering topic-specific anchor timestamps using a temporal radius of $\epsilon=12$ hours and $\mathrm{MinPts}=2$. In Eq.~\ref{eq:density}, we set the span floor to $\delta=2$ hours and the density
threshold to $\mu=0.25$ anchors per hour; each accepted interval is padded by
12 hours on both sides. Messages within these intervals are ranked using qwen3.7-text-embedding with 2,048-dimensional vectors, and the
Top-100 candidates are passed to Evidence Utility Selection. Evidence Utility Selection and
Temporal Memory Structuring both use
Deepseek-V4-Flash with thinking disabled, with temperatures of 0 and
0.2, respectively.  The evidence context is capped at $B=8192$ tokens.

We separately evaluate GPT-5.5 and Deepseek-V4-Flash as both the
answer and judge models. GPT-5.5 uses
reasoning\_effort=medium with temperature unset, whereas
Deepseek-V4-Flash uses thinking enabled and temperature 0.2.
The answer and judge output limits are 8,192 and 4,096 tokens, respectively.
Within each backbone, all methods use the same questions, answer and judge
prompts, and decoding settings. Additional configurations and retry policies
are provided in Appendix~\ref{app:settings}.

\subsection{Exp-1: Main Evaluation}
\label{sec:exp_main}

\begin{table*}[t]
\caption{Main results on GroupMemBench and EverMemBench.}
\centering

\setlength{\tabcolsep}{5pt}
\renewcommand{\arraystretch}{1.35}
\resizebox{\textwidth}{!}{%
\begin{tabular}{lccccccc cccc}
\toprule
\multirow{2}{*}{\centering\textbf{Methods}}
& \multicolumn{7}{c}{\textbf{GroupMemBench}}
& \multicolumn{4}{c}{\textbf{EverMemBench}} \\
\cmidrule(lr){2-8}\cmidrule(lr){9-12}
& \shortstack[c]{Abstention} & \shortstack[c]{Multi-\\hop}
& \shortstack[c]{Knowledge\\Update} & \shortstack[c]{Temporal}
& \shortstack[c]{Term\\Ambiguity} & \shortstack[c]{User\\Implicit}
& \shortstack[c]{Overall} & \shortstack[c]{Fine-Grained\\Recall}
& \shortstack[c]{Memory\\Awareness} & \shortstack[c]{Profile\\Understanding}
& \shortstack[c]{Overall} \\
\midrule
\rowcolor{gray!25}\multicolumn{12}{c}{GPT-5.5} \\
\midrule
Embedding & 54.68 & 54.95 & 23.36 & 55.56 & 39.62 & 55.10 & 48.32
& 43.04 & 87.78 & 41.96 & 63.25 \\
BM25 (Expand) & 62.58 & 57.69 & 20.56 & 53.08 & 38.67 & 53.06 & 49.26
& 52.89 & 84.87 & 52.31 & 67.38 \\
A-Mem & 56.12 & 46.15 & 31.78 & 35.80 & 33.01 & 61.22 & 42.82
& 41.73 & 88.14 & 50.46 & 64.92 \\
HippoRAG2 & 72.66 & 56.59 & 35.51 & 51.85 & 52.83 & 48.97 & 54.50
& 57.48 & 89.52 & 52.68 & 71.04 \\
Hindsight & 67.63 & 54.95 & 23.36 & 53.70 & 49.06 & 55.10 & 51.68
& 48.03 & 87.97 & 65.99 & 70.33 \\
StructMem & 58.27 & 55.49 & 25.23 & 58.02 & 58.49 & 57.14 & 52.75 & 51.18 & 89.33 & 58.23 & 70.21 \\
MemGAS & \textbf{74.82} & 47.25 & 33.64 & 35.80 & \textbf{59.43} & 55.10 & 50.20
& 49.08 & 88.24 & 58.78 & 69.17 \\
\method (Ours) & 70.50 & \textbf{70.88} & \textbf{39.25} & \textbf{67.28} & 55.66 & \textbf{67.35} & \textbf{63.09}
& \textbf{63.91} & \textbf{93.71} & \textbf{72.45} & \textbf{79.46} \\
\addlinespace[1pt]
\rowcolor{gray!25}\multicolumn{12}{c}{Deepseek-V4-Flash} \\
\midrule
Embedding & 53.96 & 51.10 & 34.58 & 53.70 & 34.91 & 51.02 & 47.52
& 42.78 & 86.42 & 39.74 & 62.04 \\
BM25 (Expand) & 74.10 & 48.35 & 40.19 & 55.56 & 29.25 & 46.94 & 50.74
& 50.26 & 85.23 & 50.09 & 66.21 \\
A-Mem & 54.68 & 40.10 & 31.78 & 37.65 & 28.30 & \textbf{69.39} & 41.34
& 36.48 & \textbf{91.98} & 51.20 & 65.17 \\
HippoRAG2 & 70.50 & 58.79 & 39.25 & 58.64 & 40.57 & 44.90 & 54.63
& 56.29 & 87.60 & 54.71 & 70.25 \\
Hindsight & 65.47 & 58.24 & 21.50 & 57.41 & 35.85 & 57.14 & 50.87
& 48.81 & 86.24 & 61.92 & 68.88 \\
StructMem & 57.55 & 58.24 & 26.17 & 59.88 & \textbf{52.83} & 61.22 & 53.29
& 51.97 & 89.79 & 60.25 & 71.13 \\
MemGAS & 73.38 & 40.66 & 37.38 & 32.10 & 46.23 & 55.10 & 46.17
& 46.85 & 90.51 & 53.23 & 68.25 \\
\method (Ours) & \textbf{74.82} & \textbf{70.33} & \textbf{44.86} & \textbf{69.75} & 39.62 & 65.31 & \textbf{62.68}
& \textbf{64.17} & 91.61 & \textbf{68.21} & \textbf{77.63} \\
\bottomrule
\end{tabular}%
}
\vspace{-4mm}
\label{tab:main_results}
\end{table*}

Table~\ref{tab:main_results} presents the main comparison on GroupMemBench and
EverMemBench. \method achieves the highest overall accuracy under both answer
models on both benchmarks. On GroupMemBench, it reaches 63.09\% with GPT-5.5
and 62.68\% with Deepseek-V4-Flash, improving over the strongest baseline by
8.59 and 8.05 percentage points, respectively. The advantage remains consistent
on EverMemBench, where \method exceeds the strongest completed baseline by 8.42
and 6.50 points. In the Deepseek-backed setting, StructMem is the strongest
EverMemBench baseline, reaching 71.13\% overall accuracy. These results indicate
that representing temporal structure is useful beyond a particular dataset or
answer model.

The category-level results further clarify the source of this improvement.
\method performs best on multi-hop, knowledge-update, and temporal questions
under both backbones, where answering requires either recovering distributed
evidence or distinguishing successive states. This pattern supports the joint
role of temporal evidence discovery and temporal memory structuring: the
former broadens recovery within the relevant episode, while the latter preserves
how supporting facts and state changes relate over time. \method does not dominate
every isolated category---MemGAS leads GPT-5.5 on abstention and term ambiguity,
StructMem leads Deepseek-V4-Flash on term ambiguity, and A-Mem leads
Deepseek-V4-Flash on user-implicit reasoning and EverMemBench memory awareness.
A-Mem's evolving note-level memories favor the aggregation of recurring user
attributes. StructMem's factual--relational event decomposition and cross-event
consolidation provide multiple linked cues for disambiguating recurring terms.
Nevertheless, \method's consistent overall lead
shows a better balance across retrieval, temporal reasoning, and memory
organization than either conventional retrieval or globally constructed agent
memories.

\subsection{Exp-2: Evidence Recovery}
\label{sec:exp_temporal_expansion}

\begin{wrapfigure}[15]{r}{0.43\columnwidth}
  \vspace{-4mm}
  \centering
  \includegraphics[width=0.96\linewidth]{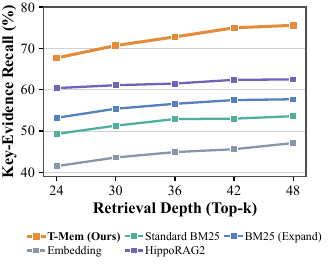}
  \vspace{-4mm}
  \caption{ Key-evidence recall across
  retrieval depths.}
  \vspace{-2mm}
  \label{fig:exp2_evidence_recovery}
\end{wrapfigure}
We test whether Temporal Evidence Discovery recovers answer-critical messages missed by global ranking on GroupMemBench questions. {We exclude Abstention questions because the absence of answer-bearing evidence is their defining property, making key-evidence recall inapplicable to these queries.}
We annotate canonical evidence offline by aligning gold answers with the original conversations and marking only messages that directly establish required answer 
claims. Evidence-level recall counts a unit as recovered if its canonical message or 
an audited, semantically equivalent alternative is retrieved; Multi-hop units
include both anchor and answer-bearing evidence. 
Gold answers and evidence labels are used only for retrospective analysis and never exposed to evaluated systems.

Figure~\ref{fig:exp2_evidence_recovery} compares global
retrieval baselines with \textsc{\method (Ours)} across Top-$k$ budgets. \method
uses BM25 (Expand) hits to locate relevant temporal regions and searches within
them; temporal memory structuring and evidence forest linearization are excluded. At Top-36, recall rises from 56.6\% with BM25 (Expand) and 61.5\%
with HippoRAG2 to 72.8\% with \method (+16.2 and +11.3 points, respectively).
Even at Top-48, HippoRAG2 reaches only 62.5\%, versus 75.6\% for \method.
The gain therefore comes from changing \emph{where} retrieval searches, not
simply from enlarging the global candidate set. High-ranked messages act as
temporal anchors that reopen nearby discussion episodes, allowing lower-ranked
messages that complete, revise, or resolve the target state to re-enter the
candidate set. This directly supports the proposed temporal-locality mechanism.
Further controlled evidence-discovery experiments and the corresponding
theoretical analysis are provided in Appendices~\ref{app:additional_results}
and~\ref{app:theory}, respectively.

\subsection{Exp-3: Memory Organization Analysis}
\label{sec:exp_forest}

We isolate organization from evidence recovery by giving both conditions the
same BM25 (Expand)-retrieved messages at each Top-$k$. \emph{BM25 (Expand)}
uses a flat chronological list, while \emph{BM25 (Expand) + Temporal Memory
Structuring} organizes the same messages as a forest and exposes
UPDATE/EXPAND relations.
Temporal Evidence Discovery and the compression step of Temporal Memory Structuring are disabled, so differences
reflect relation-based structuring alone.
Answer accuracy in this analysis is evaluated with Deepseek-V4-Flash as both
the answer and judge model.

\begin{wrapfigure}{r}{0.43\columnwidth}
  \vspace{-2mm}
  \centering
  \includegraphics[width=0.96\linewidth]{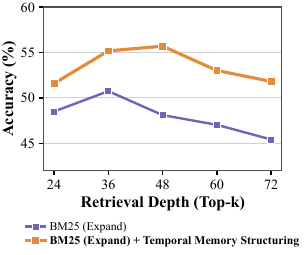}
  \vspace{-3mm}
  \caption{Accuracy with flat versus
  forest-structured evidence.}
  \vspace{-3mm}
  \label{fig:exp3_memory_organization}
\end{wrapfigure}

As shown in Figure~\ref{fig:exp3_memory_organization}, organization improves
accuracy at every retrieval depth: the gain rises from 3.09 points at Top-24 to
4.44 points at Top-36 and 7.60 points at Top-48, and remains 5.99--6.40 points
at Top-60--72. Flat BM25 (Expand) reaches 50.74\% at Top-36 and then degrades,
whereas the organized condition peaks later at 55.70\% with Top-48. Larger contexts therefore
introduce relational interference, not just more evidence: updates, obsolete
states, and supporting details become harder to distinguish in a flat list.
Explicit state-progression and support relations make additional evidence more
usable. Thus, Exp-2 and Exp-3 capture complementary mechanisms: expansion
recovers overlooked evidence, whereas the forest structures it for reasoning.
Appendix~\ref{app:additional_results} provides a finer-grained analysis of the
topological and relational contributions to this organization effect.

\begin{table*}[t]
  \caption{\textbf{Ablation study across answer/judge backbones.} Overall
  GroupMemBench accuracy (\%) across 745 questions. Values in parentheses are
  absolute changes in percentage points relative to full \method under the same
  backbone. For w/o Compression, contexts exceeding 8,192 tokens are directly
  truncated to the budget.}
  \centering
  \small
  \setlength{\tabcolsep}{12pt}
  \begin{tabular}{lcc}
    \toprule
    Method & GPT-5.5 & Deepseek-V4-Flash \\
    \midrule
    \textbf{\method (Ours)} & \textbf{63.09} & \textbf{62.68} \\
    w/o Metadata Expansion & 61.20 ($-1.89$) & 60.40 ($-2.28$) \\
    w/o Temporal Discovery & 57.05 ($-6.04$) & 56.37 ($-6.31$) \\
    w/o Temporal Structuring & 58.92 ($-4.17$) & 57.45 ($-5.23$) \\
    w/o Compression & 60.54 ($-2.55$) & 61.74 ($-0.94$) \\
    \bottomrule
  \end{tabular}
  \label{tab:ablation_models}
\end{table*}

\subsection{Exp-4: Ablation Study}
\label{sec:ablation}

We compare full \method with four controlled variants over the full GroupMemBench evaluation set, holding all other settings fixed within the GPT-5.5 and Deepseek-V4-Flash backbones. As shown in Table~\ref{tab:ablation_models}, every ablation lowers accuracy under both backbones, with reductions ranging from 0.94 to 6.31 percentage points.
Temporal Evidence Discovery provides the largest contribution, producing gains
of 6.04 and 6.31 points. This result indicates that recovering evidence from the
temporally localized episode addresses the principal retrieval bottleneck.
Removing temporal memory structuring yields the next-largest losses
(4.17 and 5.23 points), showing that the recovered messages remain less useful
when their temporal evolution and supporting relations are not organized.
Metadata expansion and compression provide smaller but consistent improvements.
Thus, every component has a positive effect across both backbones: evidence
recovery accounts for the largest improvement, temporal organization makes the
recovered evidence easier to use, and compression provides a final refinement.

\begin{figure*}[t!]
  \centering
  \vspace{-2mm}
  \includegraphics[width=\textwidth]{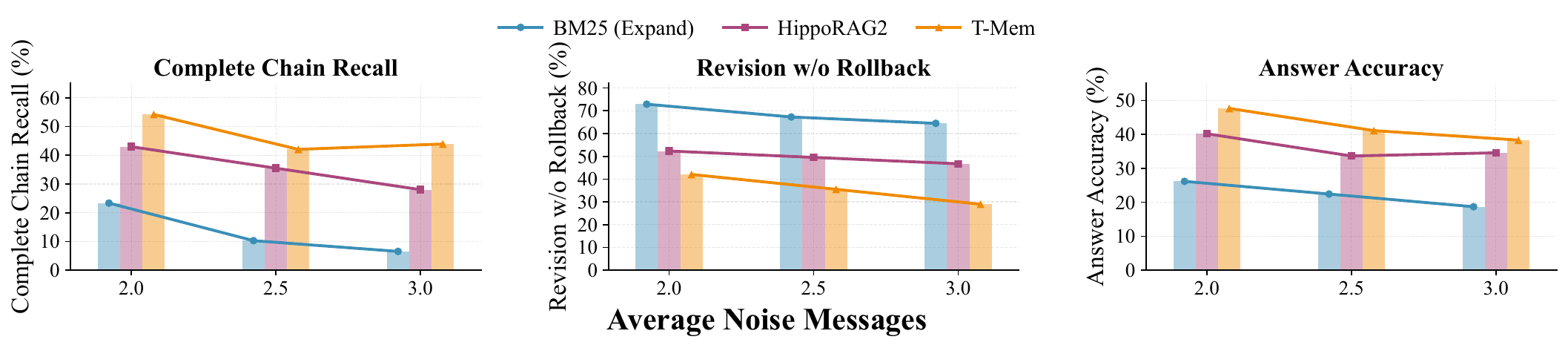}
  \vspace{-4mm}
  \caption{\textbf{Robustness to temporally local revision noise on Knowledge Update.}
  Complete Chain Recall measures recovery of all injected revisions and the final
  rollback; Revision w/o Rollback measures retrieval of intermediate revisions
  without the rollback (lower is better).}
  \vspace{-3mm}
  \label{fig:exp5_robustness}
\end{figure*}

\subsection{Exp-5: Robustness to Context Noise}
\label{sec:exp_robustness}
We augment the Knowledge Update subset with temporally local revision
chains anchored near the original evidence. Each chain introduces one or more
intermediate revisions before a final rollback restores the original state,
yielding average noise lengths of 2.0, 2.5, and 3.0 messages.
The revision chains are generated with GPT-5.6 Sol, while answers and
correctness judgments are produced with Deepseek-V4.1-Flash.
Figure~\ref{fig:exp5_robustness} evaluates three complementary outcomes.
\emph{Complete Chain Recall} measures whether all injected revisions and the
final rollback are retrieved together, indicating that the full state trajectory
is available for reasoning. \emph{Revision w/o Rollback} captures the failure
mode in which an intermediate state is retrieved but the message that invalidates
it is missed; lower is better because such partial chains favor obsolete answers.

As the chains grow, all methods face increasing interference, whereas \method
maintains stronger complete-chain recovery and fewer revision-only contexts.
Its accuracy decreases gradually from 47.66\% to 41.12\% and 38.32\%, rather
than collapsing as more intermediate states are introduced. This pattern shows
that temporal evidence discovery more reliably reconnects related state changes
and makes the rollback available alongside the revisions it supersedes. More broadly, these results demonstrate that T-Mem remains robust even as conversational structure becomes more complex.

\subsection{Exp-6: Robustness to Metadata Availability}
\label{sec:exp_metadata_robustness}

\begin{wrapfigure}[15]{r}{0.44\columnwidth}
  \vspace{-3mm}
  \centering
  \includegraphics[width=0.98\linewidth]{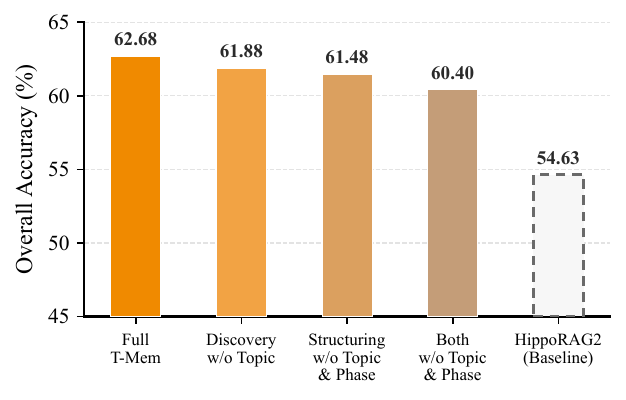}
  \vspace{-3mm}
  \caption{Overall GroupMemBench accuracy under restricted metadata
  availability. The dashed bar marks the strongest baseline.}
  \vspace{-3mm}
  \label{fig:exp6_metadata_robustness}
\end{wrapfigure}
We examine whether \method depends on the clean topic and phase annotations
available in GroupMemBench. We remove topic metadata
from Temporal Evidence Discovery, remove both topic and phase metadata from
Temporal Memory Structuring, and remove them from both modules simultaneously.
The remaining message content, timestamps, channels, authors, and roles are
unchanged. Deepseek-V4-Flash serves as both the answer and judge model, and all
other configurations match the main evaluation. Figure~\ref{fig:exp6_metadata_robustness}
also includes HippoRAG2, the strongest baseline under this backbone, as a
reference.

Restricting metadata produces a gradual rather than abrupt decline. Removing
topic from evidence discovery reduces overall accuracy by only 0.80 points,
while withholding topic and phase from structuring lowers it by 1.20 points.
Even when both modules operate without these fields, \method reaches 60.40\%,
remaining 5.77 points above HippoRAG2. The larger combined reduction indicates
that topic and phase provide useful cues for separating interleaved episodes
and interpreting their stages, but they are not the basis of the framework's
advantage. Temporal proximity still localizes candidate evidence, while message
content and chronological constraints retain enough information to reconstruct
most state transitions. The robustness therefore arises from the temporal
mechanisms themselves rather than reliance on benchmark-specific metadata.

\section{Conclusion}
\label{sec:conclusion}

In this paper, we proposed \method, a temporal-structure-based agent memory
framework that treats time as a crucial representation for managing long-term
multi-party interactions. By exploiting temporal locality to recover evidence
concentrated within relevant discussion episodes and temporal evolution to
construct relations among successive states and supporting information, \method
effectively addresses the challenges of locating critical evidence and
reconstructing complex event evolution. Experimental results on GroupMemBench
and EverMemBench demonstrate that \method consistently outperforms conventional
retrieval methods and advanced agent memory frameworks across different LLMs
and question types, while achieving favorable memory-processing efficiency.

\subsection*{AI Use Statement}

Generative AI was used for manuscript polishing and revision, as well as noise generation and data processing for a controlled experiment. Specifically, GPT-5.6 Sol generated synthetic revision-and-rollback messages following the experiment's data-construction protocol. The generation procedure is described in Appendix A.5. All methodological choices, experiments, numerical results, citations, and final claims were selected and verified by the authors.

\subsection*{Ethics Statement}

This work follows responsible research practices and uses publicly available benchmark datasets in accordance with their intended research purposes and applicable licenses. No private, proprietary, or personally identifiable group-chat data were collected or used. T-Mem is developed to support more faithful retrieval and organization of evolving conversational memories. Nevertheless, long-term memory systems may expose sensitive historical information or incorrectly surface superseded decisions when deployed in real applications. Practical deployment should therefore incorporate informed consent, access control, data minimization, retention and deletion policies, and mechanisms for auditing retrieved evidence. All reported results are intended for research evaluation rather than automated decision-making in high-stakes settings.

\subsection*{Reproducibility Statement}

Implementation details, metadata adaptation procedures, model configurations, evidence annotation methods, and prompt templates are documented in the appendices.

\bibliography{references}
\bibliographystyle{iclr2027_conference}

\appendix
\clearpage
\section{Dataset and Evidence Statistics}
\label{app:data}

\subsection{GroupMemBench}
\label{app:groupmembench}

GroupMemBench evaluates long-term memory over multi-party group conversations
in four domains: Finance, Technology, Healthcare, and Manufacturing
\citep{yang2026groupmembench}. Each domain contains 30,000 messages, giving
120,000 messages in total. The evaluation set contains 745 questions spanning
six capabilities: multi-hop reasoning, knowledge update, temporal reasoning,
user-implicit reasoning, term ambiguity, and abstention. Table~\ref{tab:gmb_stats}
reports the exact distribution used in our experiments. We retain the official
questions, gold answers, domain boundaries, and evaluation protocol.

\begin{table*}[h]
  \caption{GroupMemBench question distribution used in our experiments.}
  \centering
  \small
  \setlength{\tabcolsep}{6pt}
  \begin{tabular}{lrrrrr}
    \toprule
    Question type & Finance & Technology & Healthcare & Manufacturing & Total \\
    \midrule
    Multi-hop       & 48 & 41 & 48 & 45 & 182 \\
    Knowledge update& 32 & 36 & 17 & 22 & 107 \\
    Temporal        & 45 & 37 & 37 & 43 & 162 \\
    User implicit   & 15 & 28 &  2 &  4 &  49 \\
    Term ambiguity  & 45 & 43 &  7 & 11 & 106 \\
    Abstention      & 29 & 28 & 34 & 48 & 139 \\
    \midrule
    Total           & 214 & 213 & 145 & 173 & 745 \\
    \bottomrule
  \end{tabular}
  \label{tab:gmb_stats}
\end{table*}

\subsection{EverMemBench}
\label{app:evermembench}

EverMemBench contains long-horizon, multi-party and multi-group collaborative
dialogues together with 2,400 questions \citep{hu2026evermembench}. Its three
first-level evaluation dimensions are reported in the main comparison table;
each dimension is itself composed of three diagnostic tasks:
\begin{itemize}
  \item \textbf{Fine-grained Recall}: Single-hop retrieval tests precise local
  grounding and entity disambiguation; Multi-hop trajectory traces a person's
  work across groups; and Temporal duration identifies event boundaries and
  computes elapsed time.
  \item \textbf{Memory Awareness}: Constraint applies implicit organizational
  norms to a new situation; Proactivity detects a conflict between a new
  instruction and a stored rule; and Update resolves rule revisions and
  precedence.
  \item \textbf{Profile Understanding}: Style reproduces an individual's
  communication pattern; Skill selects actions consistent with inferred
  expertise; and Role answers from the speaker's professional perspective.
\end{itemize}
This hierarchy is important when interpreting aggregated first-level
accuracies: a single number may combine direct retrieval, cross-message
reasoning, rule application, and behavioral personalization.

\subsection{Metadata Sources and Adaptation}
\label{app:metadata_adaptation}

Table~\ref{tab:metadata_mapping} reports the exact source fields used to
construct the common message representation. GroupMemBench directly provides
the conversational metadata required by \method. EverMemBench instead stores
messages inside dated topic--group dialogue batches and supplies participant
attributes in a separate profile table. We join profiles by speaker name and
map only source-provided fields; in particular, EverMemBench has no native
phase or reply-parent annotation, so these fields are left empty rather than
inferred. The official question text and gold answer are read from
\texttt{question}/\texttt{answer} in GroupMemBench and \texttt{Q}/\texttt{A}
in EverMemBench. Gold answers and EverMemBench's \texttt{R} references are
used only for evaluation and never enter retrieval or memory construction.
For EverMemBench, the phase component is omitted from the metadata-expanded
message representation used for Seed Retrieval;
Equation~(\ref{eq:parent_candidates}) uses timestamp order within the same
topic and group without introducing a synthetic phase, stage, phase order,
or phase progress field.

\begin{table*}[t]
  \caption{Source metadata and field adaptation used for GroupMemBench and
  EverMemBench. Missing metadata is left empty rather than generated.}
  \centering
  \footnotesize
  \normalfont
  \setlength{\tabcolsep}{5pt}
  \renewcommand{\arraystretch}{1.08}
  \begin{tabular}{p{0.14\textwidth}p{0.19\textwidth}p{0.24\textwidth}p{0.34\textwidth}}
    \toprule
    Canonical field & GroupMemBench source & EverMemBench source & Adaptation and use \\
    \midrule
    Message ID & msg\_node &
      topic\_id, date, group, message\_index &
      EverMemBench has no native message-ID field; these four source fields
      form a stable provenance key that is never used as semantic evidence. \\
    Content & content & dialogue &
      Original message text retained without summarization before retrieval. \\
    Timestamp & timestamp & date, time &
      Normalized to an absolute time and used for ordering, clustering, and
      temporal constraints. \\
    Channel/group & channel (top-level JSON key) & group (key in dialogues) &
      Defines the conversational stream within which messages are compared. \\
    Phase & phase\_name & Not provided &
      Included in the metadata-expanded message representation for GroupMemBench
      and omitted for EverMemBench. \\
    Topic & topic & topic\_id &
      Defines the topic-specific anchor partition. \\
    Author/speaker & author & speaker &
      Preserves speaker identity for attribution-sensitive questions. \\
    Role/profile & role & Dept, Title, Rank (profiles) &
      EverMemBench profile fields are retained as
      auxiliary role descriptors. \\
    Ordered phase fields & stage, phase\_order, phase\_progress & Not provided &
      Used by Equation~(\ref{eq:parent_candidates}) only for GroupMemBench;
      EverMemBench uses timestamp order without phase fields. \\
    Reply parent & reply\_to & Not provided &
      Retained when available; no reply edge is synthesized for EverMemBench. \\
    \bottomrule
  \end{tabular}
  \label{tab:metadata_mapping}
\end{table*}
\subsection{Evidence Annotation and Quality for Exp-2}
\label{app:exp2_annotation}

Exp-2 evaluates whether a retrieval result contains the evidence required to
support the answer; it does not use answer accuracy as a proxy for retrieval.
Abstention questions are excluded because the absence of answer-bearing
evidence is their defining property. For each of the remaining 606 questions,
we align the query and gold answer to the original conversation and identify
one or more required \emph{evidence units}. Each required message role forms
one unit; when a question requires a cross-message bridge, its anchor and
answer-bearing message are represented as separate units. Candidate annotations
are reconstructed from the original messages using a reverse-retrieval procedure:
the gold answer is used to search the original conversation, messages related to
each key evidence claim are ranked by embedding similarity, and
Deepseek-V4-Flash filters the ranked candidates for direct evidential support.
We then manually audit low-confidence cases. Gold answers and
annotations are used only in this offline analysis and are never provided to a
retriever or answer model.

Some facts are expressed by more than one message. We therefore annotate an
alternative only when it independently establishes the same required claim;
lexical overlap or topic similarity alone is insufficient. Let
$\mathcal{U}_q$ be the required units for question $q$ and let
$\mathcal{A}(u)$ contain the canonical message for unit $u$ together with its
audited alternatives. The recall contribution of an evidence unit is
\begin{equation}
  r_q(u;R)=\mathbb{I}\!\left[R(q)\cap\mathcal{A}(u)\neq\varnothing\right].
  \label{eq:exp2_unit_recall}
\end{equation}
Thus, a unit is recovered by either its canonical message or one audited
equivalent. When an anchor is required, it forms a separate unit from the
associated answer-bearing evidence. We report micro-averaged key-evidence
recall over all questions:
\begin{equation}
  \operatorname{Recall}_{\mathrm{micro}}(R)=
  \frac{\sum_q\sum_{u\in\mathcal{U}_q} r_q(u;R)}
       {\sum_q\lvert\mathcal{U}_q\rvert}.
  \label{eq:exp2_micro_recall}
\end{equation}

\subsection{Noise Construction for Exp-5}
\label{app:noise_protocol}

Exp-5 measures whether a memory system can recover the final state after a
sequence of temporally local revisions. For each of the 107 Knowledge Update
questions, we anchor a synthetic chain near the original answer-bearing message.
The chain begins from the original state, introduces one to three successive
revisions, and ends with a rollback that restores the original state. The first
revision is highly lexically aligned with the query, later revisions have
moderate overlap, and the rollback is lexically weaker but semantically explicit.
GPT-5.6 Sol generates each chain from the question-specific original state. We
require the injected event content to remain independent of and non-duplicative
with all other information in the source conversation while preserving a
logically consistent progression from revision to rollback.
All injected messages follow the source metadata schema and remain close to the
original evidence in time. We construct three conditions with 2.0, 2.5, and 3.0
injected messages per question on average and share each fixed realization across
methods. Hidden evaluation labels identify revisions and rollbacks but are never
exposed to the retriever or answer model.

\section{Comparison Baselines}
\label{app:baselines}

We select baselines to cover distinct memory mechanisms rather than enumerate
every available implementation. \textbf{BM25} provides a transparent lexical
reference, while \textbf{Embedding} provides a dense semantic retrieval
baseline. We additionally include representative agent-memory systems:
\textbf{A-Mem} for dynamically linked and evolved memory notes,
\textbf{HippoRAG2} for graph-based associative retrieval,
\textbf{Hindsight} for memory organization through temporal- and entity-aware
networks, combining semantic, lexical, graph-based, and temporal retrieval
signals,
\textbf{StructMem} for temporally anchored event-level memories that are
periodically consolidated into semantically related cross-event
representations,
and \textbf{MemGAS} for multi-granularity memory association and selection
\citep{xu2025amem,gutierrez2025hipporag2,latimer2026hindsight,xu2026structmem,xu2026memgas}.
Together, these systems cover sparse and dense retrieval, query-independent
memory transformation, graph organization, temporal- and entity-aware memory
retrieval, temporally anchored event consolidation, and multi-granularity
selection.

This selection follows the diagnostic findings of both benchmarks. In
GroupMemBench, sophisticated memory architectures do not consistently dominate
simple lexical retrieval in long-term group conversations
\citep{yang2026groupmembench}. EverMemBench likewise reports substantial gaps
for existing memory services on multi-party attribution, temporal revision,
and implicitly relevant memories \citep{hu2026evermembench}. We therefore keep
strong simple retrieval controls rather than treating system complexity as a
proxy for quality. We do not include every service or near-duplicate memory
stack evaluated by the benchmark papers. In particular, systems whose public
interface depends on a fixed user-profile schema, omits reliable group or
timestamp propagation, or changes the official evidence interface cannot be
compared under the same retrieval--answer contract without confounding the
memory mechanism with service-specific behavior. The retained set is intended
to represent the principal design families while preserving a common
question, answer-model, and judging protocol.

\section{Detailed Experimental Settings}
\label{app:settings}

\paragraph{Model-version clarification.}
DeepSeek-V4-Flash was used for the main experiments unless otherwise specified.
Following its discontinuation from the official API, subsequent experiments used
DeepSeek-V4.1-Flash. The applicable model version is specified for each experiment;
within each controlled comparison, all compared conditions used the same answer
and judge models.

\subsection{\method Detailed Configurations}

\paragraph{Memory-processing models.}
Temporal Evidence Discovery and Temporal Memory Structuring both use
Deepseek-V4-Flash with thinking disabled. The former selects
answer-bearing evidence within retrieved episodes; the latter identifies
direct \textsc{Update} and \textsc{Expand} relations under temporal-order
constraints. Neither stage receives gold answers or annotated evidence.
Within Temporal Memory Structuring, the span-compression procedure is deterministic and
requires no additional LLM calls.

\paragraph{Retrieval and context budgets.}
Metadata-expanded BM25 retrieves the Top-36 coarse anchors, indexing message
content together with channel, phase, topic, author, and role. Queries combine
the asking-user identifier with the question; no hybrid rank fusion is used.
Within each topic, temporal clustering uses a 12-hour neighborhood and a
minimum of two anchors. Cluster density is computed as
$|C|/\max(\Delta t_C,2\text{ hours})$, with threshold 0.25 anchors per hour;
accepted intervals are padded by 12 hours on either side. The neighborhood
width controls temporal continuity, while the density threshold filters
weakly concentrated anchor groups. Within each selected interval, the Top-100
messages ranked by embedding similarity are supplied to evidence selection.
The selected evidence is merged with the coarse anchors before Temporal Memory Structuring.

The retained evidence has a token budget of $B=8192$, matching the context
limit used for the baselines. Within Temporal Memory Structuring, low-density spans are compressed
from leaves upward when this budget is exceeded, while retaining protected
spans and the forest topology. The budget concerns evidence supplied to the
answer model, not the model's output allowance; contexts already below the
limit do not require budget-driven compression.

\subsection{Baseline Configurations}

\paragraph{Models and memory construction.}
BM25 and Embedding provide direct retrieval controls without LLM-based memory
construction. The Embedding baseline uses qwen3.7-text-embedding
with 2,048-dimensional vectors and cosine similarity; BM25 uses the same
metadata expansion as \method. For A-Mem, HippoRAG2, and Hindsight, memory
construction uses Deepseek-V4-Flash, consistent with the ingestion backbone in
GroupMemBench~\citep{yang2026groupmembench}. This construction model is distinct
from the answer/judge backbones below. Our StructMem adaptation partitions
messages by channel, topic, and phase and splits each partition into
chronological chunks of at most 24 messages. It uses Deepseek-V4-Flash for
factual and relational event extraction, offline memory updating,
same-channel cross-event summarization, answering, and judging, and
qwen3.7-text-embedding with 2,048-dimensional vectors for embedding
operations. Our MemGAS adaptation uses
Deepseek-V4-Flash with thinking disabled for its LLM-based memory stages and
qwen3.7-text-embedding with 2,048-dimensional vectors for embedding operations.

A-Mem~\citep{xu2025amem} processes messages into notes with contextual
metadata and consults five neighboring memories when establishing links and
updating notes. Its implementation refreshes the retrieval index after every
100 successful memory-evolution updates, and query-time retrieval returns a
variable number of notes under the shared context cap. HippoRAG2~\citep{gutierrez2025hipporag2} performs online OpenIE
with temperature 0 and a 4,096-token extraction output limit; entity linking
uses Top-5 and final passage retrieval uses Top-36. Hindsight~\citep{latimer2026hindsight}
retains the conversation in domain-specific memory banks and uses native
recall with \texttt{budget=mid} and \texttt{max\_tokens=8192}. Its returned
memory count is variable: no additional Top-1 or fixed Top-$k$ cutoff is
applied. StructMem~\citep{xu2026structmem} runs in event mode and extracts both
factual and relational memories. It merges sufficiently similar updates at an
offline cosine threshold of 0.90 and forms same-channel cross-event summaries
using a one-hour temporal window and 15 semantic seeds. At query time, it
retrieves the Top-60 atomic memories and Top-5 cross-event summaries by cosine
similarity before enforcing the shared 8,192-token context cap.
MemGAS~\citep{xu2026memgas}
constructs session-, turn-, summary-, and keyword-level memories, retrieves the
Top-36 sessions, and applies its multi-granularity association and selection
procedure before answer generation.

\paragraph{Retrieval limits.}
The seven baselines share an 8,192-token evidence-context cap, but retain their
respective retrieval units and depths, as summarized in
Table~\ref{tab:baseline_retrieval_limits}. A Top-$k$ limit bounds the number
of returned items, whereas the token cap bounds their combined length.
Consequently, short retrieval results remain unchanged; truncation is applied
only when the returned context exceeds the cap.
For w/o Compression, contexts exceeding 8,192 tokens are directly truncated
to the budget.

\begin{table}[htbp]
\caption{Baseline retrieval limits. Context caps apply to retrieved evidence, not to generated answers or judge outputs.}
\centering
\small
\setlength{\tabcolsep}{10pt}
\begin{tabular}{lcc}
\toprule
Method & Retrieval Top-$k$ & Context cap (tokens) \\
\midrule
Embedding & 36 & 8,192 \\
BM25 & 36 & 8,192 \\
A-Mem & Variable & 8,192 \\
HippoRAG2 & 36 & 8,192 \\
Hindsight & Variable & 8,192 \\
StructMem & 60 atomic + 5 summaries & 8,192 \\
MemGAS & 36 & 8,192 \\
\bottomrule
\end{tabular}

\label{tab:baseline_retrieval_limits}
\end{table}

\subsection{Answer and Judge Model Settings}

We evaluate two answer/judge backbones: GPT-5.5 and Deepseek-V4-Flash.
For each backbone, the same model generates answers and judges them in
separate calls. GPT-5.5 uses reasoning\_effort=medium;
Deepseek-V4-Flash uses thinking enabled. The standard output allowances are
8,192 tokens for answering and 4,096 tokens for judging, separately from the
8,192-token evidence-context budget. These allowances accommodate reasoning
as well as the required final response.

Within each controlled comparison, methods use the same question set,
answer/judge backbone, common answer requirements, and judging protocol.
Evidence presentation remains method-specific: flat retrieval baselines
provide passages, whereas \method exposes the temporal evidence forest. The judge
receives the question, reference answer, and generated answer, accepts
meaning-equivalent paraphrases, and returns \texttt{Final: Correct} or
\texttt{Final: Incorrect}. Reference answers are never supplied to retrieval
or answer generation. For the Embedding evaluation with
Deepseek-V4-Flash, an empty output
triggers a retry with thinking disabled; completed questions are skipped
rather than evaluated again. This is an output-recovery exception, not a
change to the retrieved evidence.

\section{Reproducibility}
\label{app:reproducibility}

\subsection{Stage-wise Execution}

The implementation separates coarse retrieval, temporal expansion, forest
organization, compression, answer generation, and judging into explicit stages.
All questions finish one stage before the next stage begins. Each stage writes
append-only JSONL records containing the stable question key, configuration,
selected message identifiers, model outputs, status, and errors. This design
allows intermediate evidence sets and forests to be inspected independently
of downstream accuracy.

\subsection{Resume and Failure Handling}

Each runner reconstructs the latest record for the tuple
\texttt{(domain, question\_type, question\_id)}. Successful records are skipped
on restart, while failed or incomplete records are retried. Document
embeddings, query embeddings, temporal-window selections, and completed forest
records are cached separately so that a transient model-service or cluster
failure does not require repeating earlier stages. Structured LLM calls use
bounded retries for timeouts, empty responses, and malformed JSON; persistent
failures remain visible as error records rather than being silently omitted.

\subsection{Configuration and Result Integrity}

Run contracts store model names, retrieval depths, embedding dimensions,
clustering parameters, prompt hashes, and output directories. Validators check
the official question counts, uniqueness of question keys, temporal ordering,
single-parent forests, and the absence of gold answers from retrieval and
answer prompts. Prompt and package manifests are hashed before execution.
Reported summaries are generated from the latest successful record per stable
key, while raw JSONL histories are retained for audit.

\section{Additional Results}
\label{app:additional_results}

\subsection{Efficiency Measurements}
\label{app:efficiency}

Table~\ref{tab:groupmembench_efficiency} evaluates efficiency primarily through
total token consumption, which directly reflects the amount of model processing
required by each method. Monetary cost is reported as a complementary indicator,
since it also depends on the prices of the models used at different stages.
Memory-processing tokens cover corpus construction or retrieval, whereas
answer-and-judge tokens cover the downstream evaluation. For Embedding, the
token count includes one encoding pass over the 120,000-message corpus and all
745 queries; because the encoder can be deployed locally, this processing does
not incur an embedding API charge. BM25 requires no model-based memory
processing.

\begin{table*}[t]
\caption{\textbf{Efficiency on GroupMemBench.} Token consumption is the
primary efficiency measure, while total API-equivalent cost is reported as a
complementary indicator for the complete 745-question evaluation. Embedding
tokens are counted although the encoder can be deployed locally. The StructMem
and MemGAS costs are estimated from stage-level token usage; downstream answer
and judge calls use GPT-5.5-equivalent input/output pricing.}
\centering
\small
\setlength{\tabcolsep}{7pt}
\renewcommand{\arraystretch}{1.12}
\begin{tabular}{@{}lccccc@{}}
\toprule
Method &
\shortstack{Memory Processing\\Tokens} &
\shortstack{Answer and Judge\\Tokens} &
\shortstack{Total\\Tokens} &
\shortstack{Average Tokens\\per Question} &
\shortstack{Total Cost\\(USD)} \\
\midrule
Embedding        & 9.07M   & 5.80M & 14.87M  & 19.96K  & 39.64 \\
BM25 (Expand)     & 0       & 5.66M & 5.66M   & 7.60K   & 38.68 \\
A-Mem             & 392.08M & 1.11M & 393.19M & 527.77K & 76.70 \\
HippoRAG2          & 129.57M & 5.47M & 135.04M & 181.26K & 65.50 \\
Hindsight         & 481.67M & 5.68M & 487.35M & 654.16K & 122.59 \\
StructMem         & 125.16M & 8.33M & 133.49M & 179.18K & 148.67 \\
MemGAS            & 35.63M  & 5.47M & 41.10M  & 55.17K  & 46.84 \\
\textbf{\method (Ours)} & \textbf{26.43M} & \textbf{5.15M} &
\textbf{31.58M} & \textbf{42.39K} & \textbf{44.04} \\
\bottomrule
\end{tabular}

\label{tab:groupmembench_efficiency}
\end{table*}

Compared with conventional retrieval-augmented memory methods, \method consumes moderately more tokens: 31.58M in total, compared with 14.87M for Embedding and 5.66M for BM25. This additional computation supports temporal evidence recovery and structured memory organization. Nevertheless, these memory-processing stages use a
comparatively inexpensive model, so the monetary difference remains limited:
\method costs USD~44.04, only USD~4.40 more than Embedding and USD~5.36 more than BM25.

The contrast is larger for methods that use LLMs to construct or select
global memory representations. MemGAS, StructMem, HippoRAG2, A-Mem, and Hindsight
consume 41.10M, 133.49M, 135.04M, 393.19M, and 487.35M tokens, respectively,
whereas \method requires only 31.58M. Thus, \method reduces total token
consumption by 23.2\%, 76.3\%, 76.6\%, 92.0\%, and 93.5\% relative to these five
baselines. Its total cost is also lower, at USD~44.04 versus USD~46.84,
USD~148.67, USD~65.50, USD~76.70, and USD~122.59, respectively. These results show that
query-conditioned temporal processing avoids the high cost of globally
constructing and repeatedly consolidating LLM-derived memories. Overall, \method
offers a more efficient balance: it adds only moderate processing over
traditional retrieval, yet requires far fewer tokens than existing agent memory
frameworks while retaining their structured-memory capability.

\subsection{Evidence Discovery Analysis}
\label{app:evidence_discovery_analysis}

Exp-2 shows that Temporal Evidence Discovery can recover answer-critical
messages missed by global retrieval. However, its expanded memory set may
contain more messages than the initial retrieval result, so that comparison
alone cannot isolate temporal locality from a larger candidate or output
budget. We therefore introduce two fixed-budget variants. \emph{T-Mem
Reranked} applies Qwen3-Reranker-8B to the candidates recovered by Temporal
Evidence Discovery and retains exactly 36 messages. \emph{Global Evidence
Retrieval} uses the same seed results and topic partition but removes temporal
localization: if \method selects \(n\) temporal clusters for a topic, the
control instead takes the embedding Top-\(100n\) candidates from the entire
topic, applies the same Evidence Utility Selection, and reranks the survivors
to Top-36. Both variants use qwen3.7-text-embedding with 2,048-dimensional
vectors, Deepseek-V4-Flash for Evidence Utility Selection under the same
decoding configuration, and Qwen3-Reranker-8B for final reranking. All other
retrieval parameters are held fixed; only the candidate search region differs.

\begin{figure*}[t]
  \centering
  \includegraphics[width=0.459\textwidth,height=0.344\textwidth]{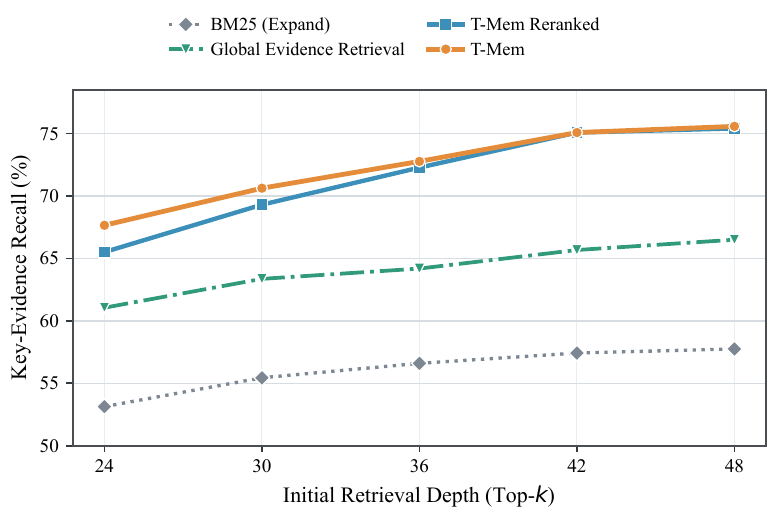}%
  \hfill
  \includegraphics[width=0.495\textwidth,height=0.344\textwidth]{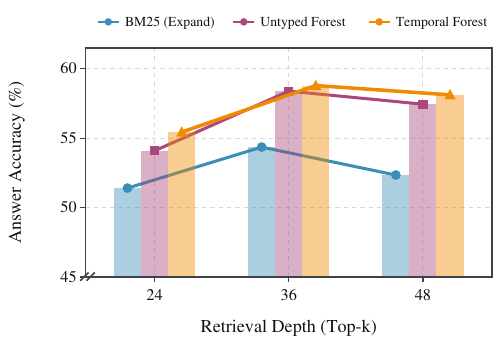}
  \caption{Analysis of Temporal Evidence Discovery and Temporal Memory
  Structuring. \textbf{Left:} key-evidence recall over 606 non-Abstention
  questions. T-Mem Reranked and Global Evidence Retrieval both output exactly
  36 messages and share the same embedding, evidence-selection, and reranking
  models and configurations; the global control searches the entire topic
  instead of temporally localized intervals. The horizontal axis denotes the
  initial retrieval depth.
  \textbf{Right:} answer accuracy under flat, untyped-forest, and typed
  temporal-forest representations at different retrieval depths, using
  DeepSeek-V4.1-Flash as the answer and judge model.}
  \label{fig:fixed_budget_evidence}
\end{figure*}

Figure~\ref{fig:fixed_budget_evidence} shows that controlling the final context
size preserves nearly all of the discovery gain: at an initial depth of 36,
T-Mem Reranked reaches 72.3\% recall, only 0.5 percentage points below
unrestricted \method. More importantly, T-Mem Reranked consistently exceeds
Global Evidence Retrieval even though the two controls share the same
candidate capacity, final Top-36 budget, retrieval models, and decoding
settings. At the same initial depth of 36, temporal localization raises recall
from 64.2\% to 72.3\%, and the advantage persists across all tested depths.
Since the controlled variants differ only in whether dense candidate search is
restricted to the query-induced temporal intervals, the gain cannot be
attributed to returning more messages, stronger models, or an additional
reranking stage. Instead, the localized intervals concentrate retrieval on the
relevant episode and expose answer-critical evidence that remains low-ranked
within the complete topic. This provides direct controlled evidence that the
benefit of Temporal Evidence Discovery arises from Temporal Locality rather
than global candidate expansion.

\subsection{Relation Semantics Analysis}
\label{app:relation_semantics_analysis}

Exp-3 compares a flat list with the complete Temporal Memory Structuring module
under an identical retrieved message set. This establishes the benefit of
organization, but combines the effect of the forest topology with the semantics
of its relation labels. To separate these factors, we add an untyped-forest
control. At each Top-$k$, it preserves the parent assignments and topology of
the BM25 (Expand) + Temporal Memory Structuring forest, while replacing every
\textsc{Update} and \textsc{Expand} label with a single \textsc{Related}
relation. All three representations use the same retrieved messages, with
DeepSeek-V4.1-Flash serving as both the answer and judge model.

The right panel of Figure~\ref{fig:fixed_budget_evidence} shows that both forest
representations consistently outperform the flat list, while the typed temporal
forest achieves the highest accuracy at every retrieval depth. The untyped
forest already yields a clear improvement, indicating that parent--child
topology helps the model follow connected evidence rather than interpret each
message independently. Restoring \textsc{Update} and \textsc{Expand} semantics
provides a further, consistent gain; at Top-36, for example, the typed forest
reaches 58.79\% accuracy, compared with 58.39\% for the untyped forest and
54.36\% for the flat list. The additional benefit is most pronounced when the
retrieval budget is smaller, where relation labels help distinguish
state-changing revisions from supporting elaborations despite limited context.
These results indicate that the gain from Temporal Memory Structuring is driven
primarily by organizing messages into an event progression, with typed
relations further clarifying how each memory changes or supports that
progression.

\clearpage
\section{Theoretical Analysis}
\label{app:theory}

\subsection{Reliability of Density-Based Episode Selection}
\label{app:theory_expansion}

We provide an idealized analysis of the density rule in
Eq.~\ref{eq:density}. To avoid cumbersome nested subscripts, $C$ below denotes
the same topic-specific anchor cluster written as $C_j$ in the main text.
Consider its induced candidate interval with observable duration
$\Delta_C\geq\delta$, and let $A_C=\lvert C\rvert$ be its number of anchors.
We distinguish a relevant episode ($H_1$) from background activity ($H_0$).

\paragraph{Assumption 1 (Anchor-rate separation).}
Conditional on $H_h$, anchor arrivals within $C$ follow a homogeneous Poisson
process with rate $\lambda_h$, where $0<\lambda_0<\lambda_1$. Hence,
$A_C\mid H_h\sim\operatorname{Poisson}(\lambda_h\Delta_C)$. This assumption
formalizes the condition motivating Temporal Evidence Discovery: anchors are
more concentrated in answer-relevant episodes than in background intervals.

\paragraph{Proposition 1 (Density form of the likelihood-ratio test).}
Under Assumption~1, equal class priors, and equal error costs, the
likelihood-ratio test selects $H_1$ exactly when
$A_C/\Delta_C>\mu^\star$, where
\begin{equation}
  \mu^\star=
  \frac{\lambda_1-\lambda_0}
       {\log(\lambda_1/\lambda_0)},
  \qquad \lambda_0<\mu^\star<\lambda_1.
  \label{eq:optimal_density_threshold}
\end{equation}
\textit{Proof.}
For an observed count $a$, the log-likelihood ratio is
\begin{equation}
  \log\frac{\Pr(A_C=a\mid H_1)}{\Pr(A_C=a\mid H_0)}
  =a\log\frac{\lambda_1}{\lambda_0}
   -(\lambda_1-\lambda_0)\Delta_C.
  \label{eq:poisson_llr}
\end{equation}
It is positive precisely when $a/\Delta_C>\mu^\star$. The strict inequalities
in Eq.~\ref{eq:optimal_density_threshold} follow from the mean value theorem
applied to $\log x$. \hfill$\square$

\paragraph{Theorem 1 (Exponential decay of selection errors).}
Choose any density threshold $\mu$ satisfying
$\lambda_0<\mu<\lambda_1$, and define
\begin{equation}
  D_{\mathrm P}(a\Vert b)
  =a\log\frac{a}{b}-a+b.
  \label{eq:poisson_divergence}
\end{equation}
Then the false-positive and false-negative probabilities obey
\begin{align}
  \Pr\!\left(\frac{A_C}{\Delta_C}\geq\mu\,\middle|\,H_0\right)
  &\leq
  \exp\!\left[-\Delta_C D_{\mathrm P}(\mu\Vert\lambda_0)\right],
  \label{eq:density_fp_bound}\\
  \Pr\!\left(\frac{A_C}{\Delta_C}\leq\mu\,\middle|\,H_1\right)
  &\leq
  \exp\!\left[-\Delta_C D_{\mathrm P}(\mu\Vert\lambda_1)\right].
  \label{eq:density_fn_bound}
\end{align}
\textit{Proof.}
If $A\sim\operatorname{Poisson}(\lambda\Delta)$, its moment-generating
function is
$\mathbb{E}[e^{tA}]=\exp\{\lambda\Delta(e^t-1)\}$.
Applying the Chernoff bound to the upper tail and minimizing over $t>0$
gives $t^\star=\log(\mu/\lambda_0)$, which yields
Eq.~\ref{eq:density_fp_bound}. Applying the same argument to the lower tail
with $t<0$ gives $t^\star=\log(\mu/\lambda_1)$ and
Eq.~\ref{eq:density_fn_bound}. Because
$D_{\mathrm P}(a\Vert b)>0$ whenever $a\neq b$, both exponents are strictly
negative. \hfill$\square$

\paragraph{Conclusion.}
Under separated anchor rates, both kinds of density-threshold error therefore
decrease exponentially with the observable interval length. This result
justifies using anchor density rather than anchor count alone. For
$\Delta_C<\delta$, the denominator $\max(\Delta_C,\delta)$ is a short-interval
regularizer; the exponential statement above is made only for
$\Delta_C\geq\delta$.

\paragraph{Scope of the result.}
The guarantee is conditional on Assumption~1, an interval treated as fixed
before its tested anchor count is observed, and a threshold strictly between
the two anchor rates. It establishes the
reliability of the density decision under this model; it does not guarantee
that the interval generator recovers every relevant episode, that real anchor
arrivals are exactly homogeneous Poisson, or that selected evidence is used
correctly by the answer model. In practice, candidate intervals may be
determined by a clustering algorithm or a learned selector.

\subsection{Justification of Compression in Temporal Memory Structuring}
\label{app:theory_compression}

Let $\mathcal{S}$ contain all spans considered for compression, and let $p_s$
denote the parent of the leaf containing span $s$. Let $\mathcal{M}_q$ contain
all mandatory spans: protected query entities, timestamps, numerical values,
state-changing content, and one retained span per node.

\paragraph{Assumption 2 (Budget feasibility).}
The mandatory content fits the answer-time budget:
\begin{equation}
  \sum_{s\in\mathcal{M}_q}\ell(s)\leq B.
  \label{eq:mandatory_budget}
\end{equation}
Without this condition, no compression rule that preserves every protected
item and every node can satisfy the budget.

\paragraph{Lemma 1 (Smoothed IDF as self-information).}
For lexical unit $u$, define its smoothed empirical document-occurrence
probability by
\begin{equation}
  \widehat{P}(u)=\frac{\operatorname{df}(u)+1}{N+1}.
\end{equation}
Its self-information is
\begin{equation}
  -\log\widehat{P}(u)
  =\log\frac{N+1}{\operatorname{df}(u)+1}.
  \label{eq:idf_self_information}
\end{equation}
Thus, the weight in Eq.~\ref{eq:information_gain} is a smoothed
self-information term with a unit floor. Rare units contribute more, while the
query-overlap and parent-novelty sets specify which informative units are
useful in context. \hfill$\square$

Consider the budgeted span-selection problem
\begin{equation}
  \begin{aligned}
  \max_{\{x_s\}}\quad
    &\sum_{s\in\mathcal{S}}x_s G_q(s\mid p_s)\\
  \text{s.t.}\quad
    &\sum_{s\in\mathcal{S}}x_s\ell(s)\leq B,\\
    &x_s=1 \quad \forall s\in\mathcal{M}_q,\\
    &x_s\in\{0,1\} \quad \forall s\notin\mathcal{M}_q.
  \end{aligned}
  \label{eq:discrete_compression_problem}
\end{equation}
Here, $x_s=1$ means that span $s$ is retained. The objective preserves as much
query-conditioned information as possible under the token budget.

\paragraph{Proposition 2 (Optimality for the continuous relaxation).}
Relaxing the optional variables in
Eq.~\ref{eq:discrete_compression_problem} to $x_s\in[0,1]$ admits an optimal
solution that retains optional spans in nonincreasing order of
\begin{equation}
  g_q(s\mid p_s)=\frac{G_q(s\mid p_s)}{\ell(s)},
\end{equation}
with at most one fractional boundary span after ties are resolved.

\textit{Proof.}
Let
$B_q=B-\sum_{s\in\mathcal{M}_q}\ell(s)\geq0$ be the residual budget after
mandatory spans are fixed. For the optional spans, the Lagrangian of the
relaxed problem is
\begin{equation}
  \mathcal{L}(x,\eta)
  =\eta B_q+
   \sum_{s\notin\mathcal{M}_q}
   x_s\ell(s)\bigl(g_q(s\mid p_s)-\eta\bigr),
  \qquad \eta\geq0.
  \label{eq:compression_lagrangian}
\end{equation}
For a fixed multiplier $\eta$, separability makes $x_s=1$ optimal when
$g_q(s\mid p_s)>\eta$ and $x_s=0$ optimal when
$g_q(s\mid p_s)<\eta$; only spans at the boundary may be fractional. Choosing
$\eta$ to satisfy the residual budget therefore yields the stated density
ordering. \hfill$\square$

\paragraph{Conclusion.}
Compressing nonmandatory spans from low to high information density is optimal
for the continuous relaxation of the token-budget problem. Because the
implemented procedure retains or omits whole spans, the discrete problem is a
0--1 knapsack problem; the same ordering is therefore a computationally
efficient approximation rather than a claim of exact discrete optimality.
Node metadata, edges, mandatory evidence, and at least one span per node remain
unchanged.

\clearpage
\section{Examples and Case Studies}
\label{app:cases}

We provide qualitative examples to complement the controlled mechanism
analyses in the main paper. These examples illustrate how \method's two temporal mechanisms operate
on individual instances and are not intended as additional causal evidence. We first
examine a case in which global retrieval identifies relevant messages
but misses the message that determines the current state.

\subsection{Recovering an Overlooked State Update}
\label{app:case_temporal_discovery}

Figure~\ref{fig:case_temporal_discovery} illustrates the temporal-locality
mechanism analyzed in the evidence-recovery experiment. The query asks for
the \emph{current} approach to reviewing channel inconsistencies. Although
global retrieval identifies messages from the relevant discussion, both
BM25 (Top-36) and HippoRAG2 (Top-36) miss the decisive state-changing message
\texttt{Msg\_6416}. Temporal Evidence Discovery instead uses the retrieved
anchors to revisit the relevant discussion episode and
recovers the overlooked update. The recovered message explicitly supersedes
the earlier ``core first, assisted later'' policy and specifies that the
assisted channel should be reviewed together with web, email, and in-app
during the first pass.

\begin{figure}[!ht]
    \centering

    \begin{casebox}{
        Temporal Evidence Discovery Recovers a Missed State Update
    }

    \footnotesize
    \textbf{Instance.}
    \texttt{Finance / knowledge\_update\_27}
    \vspace{1.5mm}
    \textbf{Question.}

    \emph{What is the team's current approach for reviewing channel
    inconsistencies in the customer journey?}

    \vspace{2mm}
    \hrule
    \vspace{2mm}

    \textbf{Global retrieval}

    \vspace{1mm}

    \textcolor{casepink}{\textbf{BM25 (Top-36)}}:
    retrieves messages from the relevant discussion, but
    \textcolor{casepink}{
        \textbf{does not retrieve \texttt{Msg\_6416}}
    }.

    \vspace{1mm}

    \textcolor{casepink}{\textbf{HippoRAG2 (Top-36)}}:
    also misses \texttt{Msg\_6416}; its answer does not reflect the
    updated review policy.

    \hfill
    \textcolor{casepink}{\textbf{Incorrect}}

    \vspace{2.5mm}

    \textbf{Temporal Evidence Discovery}

    \vspace{1mm}

    Retrieved anchors identify the relevant discussion episode. Searching within this local region recovers the
    lower-ranked but state-changing message
    \textcolor{caseblue}{\textbf{\texttt{Msg\_6416}}}.

    \vspace{1.5mm}

    \begin{evidencebox}

    \textbf{\texttt{Msg\_6416} --- recovered key evidence}

    \vspace{0.8mm}

    The original ``core first, assisted later'' decision no longer holds
    because assisted matches the trigger and fallback wording, yet its owner
    handoff and last-edit history diverge.

    \vspace{0.8mm}

    \textcolor{caseblue}{
        \textbf{New plan:}
        include assisted in the first pass with web/email/in-app, and check
        all four together for owner drift and handoff inconsistency.
    }

    \end{evidencebox}

    \vspace{2mm}

    \textbf{Gold answer}

    \vspace{0.8mm}

    Instead of keeping assisted as a follow-up, include assisted in the
    first pass so all four channels are checked together for owner drift
    and handoff inconsistencies.

    \vspace{2mm}
    \textbf{\method answer (excerpt)}

    \vspace{0.8mm}

    The team's current approach is to lock a single unified/core onboarding
    journey, allow only light channel-specific presentation differences,
    and have Support validate
    \textcolor{caseblue}{
        \textbf{web, email, in-app, and assisted}
    }
    paths against the spec.

    \hfill
    \textcolor{casegreen}{\textbf{Correct}}

    \end{casebox}

    \caption{
    \textbf{Qualitative example of Temporal Evidence Discovery.}
    BM25 and HippoRAG2 at Top-36 retrieve relevant messages but miss the
    state-changing message \texttt{Msg\_6416}. Temporal locality allows
    \method to revisit the corresponding discussion episode, recover the
    overlooked update, and reconstruct the current four-channel review
    policy. Blue text highlights the answer-critical information recovered
    from the local temporal episode.
    }

    \label{fig:case_temporal_discovery}

\end{figure}

The case highlights a distinction between \emph{topic relevance} and
\emph{state completeness}. Global retrieval can identify the correct topic
while still omitting a lower-ranked message that revises the state relevant
to the query. In this instance, the decisive information is not a new topic
but a local update that changes the scope of the existing policy. Temporal
Evidence Discovery therefore recovers missing evidence by changing
\emph{where} the system searches after coarse retrieval, rather than merely
increasing the global retrieval depth. This example also shows why a
current-state question is sensitive to a single revision even when many
topically relevant messages are already available. Recovering
\texttt{Msg\_6416} changes the interpretation of the surrounding evidence
from a staged review process to a unified first-pass policy. The case thus
illustrates how temporal locality complements similarity ranking by restoring
state-changing evidence that would otherwise remain outside the retrieved
context.

\begin{figure}[!ht]
    \centering

    \begin{casebox}{
        Temporal Memory Structuring Tracks an Evolving Audit-Retention Decision
    }

    \footnotesize

    \textbf{Instance.}
    \texttt{Manufacturing / knowledge\_update\_3}

    \vspace{1.5mm}

    \textbf{Question.}

    \emph{What is the current approach for ensuring compliance with
    audit retention requirements in the training rollout?}

    \vspace{2mm}
    \hrule
    \vspace{2mm}

    \textbf{Flat evidence presentation}

    \vspace{1mm}

    The retrieved messages contain both the earlier LMS-export approach and
    later concerns about immutable audit records. Presented as a flat list,
    these states are not explicitly distinguished as successive revisions.

    \vspace{1mm}

    \textcolor{casepink}{\textbf{BM25 (Top-36) answer}}:
    emphasizes aligning LMS export fields and notes the concern that LMS
    exports alone may not satisfy Legal's retention requirement, but does
    not identify the proposed append-only logging approach or the required
    schema extension before the L\&D build.

    \hfill
    \textcolor{casepink}{\textbf{Incorrect}}

    \vspace{2.5mm}

    \textbf{Temporal Memory Structuring}

    \vspace{1mm}

    \method links three messages from this discussion into an explicit
    sequence of state updates:

    \vspace{1.5mm}

    \begin{evidencebox}

    \textbf{\texttt{Msg\_3433} --- earlier state}

    \vspace{0.8mm}

    Retention fields match the LMS export schema; HRIS
    \texttt{cert\_status} and \texttt{completion\_date} are confirmed.

    \end{evidencebox}

    \vspace{0.8mm}

    \begin{center}
        \textcolor{caseblue}{\textbf{\textsc{Update} $\Downarrow$}}
    \end{center}

    \vspace{0.3mm}

    \begin{evidencebox}

    \textbf{\texttt{Msg\_5244} --- challenge to the earlier approach}

    \vspace{0.8mm}

    LMS exports alone may not satisfy Legal's retention requirement because
    auditors expect an immutable completion history.

    \end{evidencebox}

    \vspace{0.8mm}

    \begin{center}
        \textcolor{caseblue}{\textbf{\textsc{Update} $\Downarrow$}}
    \end{center}

    \vspace{0.3mm}

    \begin{evidencebox}

    \textbf{\texttt{Msg\_6194} --- current state}

    \vspace{0.8mm}

    The LMS export approach may not meet the immutable-audit requirement.

    \vspace{0.8mm}

    \textcolor{caseblue}{
        \textbf{New plan:}
        add an append-only completion log
        (LRS/xAPI or an immutable audit table) and extend the schema
        before the L\&D build proceeds.
    }

    \end{evidencebox}

    \vspace{2mm}

    \textbf{Gold answer}

    \vspace{0.8mm}

    Instead of relying solely on LMS export fields, implement an append-only
    completion log, such as LRS/xAPI or an immutable audit table, and extend
    the schema before the L\&D build proceeds.

    \vspace{2mm}

    \textbf{\method answer (excerpt)}

    \vspace{0.8mm}

    The current approach is to move beyond LMS exports and add an
    \textcolor{caseblue}{
        \textbf{append-only completion history log}
    },
    either through LRS/xAPI or an immutable audit table, before the
    L\&D build proceeds.

    \hfill
    \textcolor{casegreen}{\textbf{Correct}}

    \end{casebox}

    \caption{
    \textbf{Qualitative example of Temporal Memory Structuring.}
    An explicit \textsc{Update} chain distinguishes the earlier LMS-export
    policy, its challenge, and the revised append-only completion plan.
    Blue text marks the current update; content is condensed for readability.
    }

    \label{fig:case_temporal_structuring}

\end{figure}
\subsection{Organizing Successive States with Temporal Relations}
\label{app:case_temporal_structuring}

Figure~\ref{fig:case_temporal_structuring} illustrates how Temporal Memory
Structuring represents the evolution of an audit-retention decision.
In this example, the discussion initially relies on the LMS export schema
for audit retention. A later message
questions whether these exports satisfy the immutable-record requirement,
and the latest update proposes adding an append-only completion log
and extending the schema before the L\&D build proceeds. Temporal Memory Structuring links these messages through successive
\textsc{Update} relations to present the sequence of revisions to the
answer model.

This example illustrates how temporal relations represent successive
revisions within a discussion. The temporal evidence forest links the earlier
LMS-export approach, the subsequent challenge, and the proposed revision as
\texttt{Msg\_3433 $\rightarrow$ Msg\_5244 $\rightarrow$ Msg\_6194} through
\textsc{Update} edges. It complements the controlled memory-organization
analysis in the main paper by showing how the inferred relation structure
is presented to the answer model. The displayed answers alone do not
establish a causal effect of structuring. The important distinction is that
all three messages remain relevant to the topic, but they do not represent
equivalent states of the decision. The explicit \textsc{Update} chain makes
the direction of revision visible and prevents the earlier LMS-only approach
from being treated as co-current with the later proposal. This organization
therefore provides the answer model with a compact representation of both
decision history and state precedence.

\subsection{Recovering a Newly Added Field Constraint}
\label{app:case_field_constraint}

Figure~\ref{fig:case_field_constraint} provides a second example of how
Temporal Evidence Discovery recovers a low-ranked state update that is
critical for answering a current-state query. The discussion contains many
messages about field tagging and fallback behavior, allowing global retrieval
to reach the correct topic. However, both BM25 (Top-36) and HippoRAG2
(Top-36) fail to retrieve \texttt{Msg\_6579}, which introduces an additional
fallback-eligibility constraint. Temporal Evidence Discovery revisits the
corresponding local discussion episode and recovers this update, allowing
\method to reconstruct the complete two-part tagging rule.

\begin{figure}[t]
    \centering

    \begin{casebox}{
        Temporal Evidence Discovery Recovers a Newly Added Field Constraint
    }

    \footnotesize

    \textbf{Instance.}
    \texttt{Technology / knowledge\_update\_31}

    \vspace{1.5mm}

    \textbf{Question.}

    \emph{What is the current approach for tagging fields in the data
    integration setup?}

    \vspace{2mm}
    \hrule
    \vspace{2mm}

    \textbf{Global retrieval}

    \vspace{1mm}

    \textcolor{casepink}{\textbf{BM25 (Top-36)}}:
    retrieves messages about field states and integration rules, but
    \textcolor{casepink}{
        \textbf{does not retrieve \texttt{Msg\_6579}}
    }.

    \vspace{1mm}

    \textcolor{casepink}{\textbf{HippoRAG2 (Top-36)}}:
    also misses \texttt{Msg\_6579} and remains focused on the earlier
    three-state tagging scheme without recovering the newly added
    fallback-eligibility constraint.

    \hfill
    \textcolor{casepink}{\textbf{Incorrect}}

    \vspace{2.5mm}

    \textbf{Temporal Evidence Discovery}

    \vspace{1mm}

    Retrieved anchors locate the relevant field-tagging discussion episode.
    Searching within this local temporal region recovers the lower-ranked
    update
    \textcolor{caseblue}{\textbf{\texttt{Msg\_6579}}}.

    \vspace{1.5mm}

    \begin{evidencebox}

    \textbf{\texttt{Msg\_6579} --- recovered key evidence}

    \vspace{0.8mm}

    \textcolor{caseblue}{
        \textbf{New rule:}
        keep \textbf{stable / blocked-for-mapping / unstable-but-visible}
        for display, but add a per-field
        \textbf{fallback-eligibility} flag:
        \textbf{safe-copy} vs.\ \textbf{block/hide}.
    }

    \end{evidencebox}

    \vspace{2mm}

    \textbf{Gold answer}

    \vspace{0.8mm}

    Instead of relying on the three-state tag alone, use a two-part rule:
    retain the state tag for visibility while adding an explicit
    fallback-eligibility flag so that reporting can distinguish fields that
    may be shown with safe copy from those that should be hidden or blocked.

    \vspace{2mm}

    \textbf{\method answer (excerpt)}

    \vspace{0.8mm}

    The current approach keeps the three display states
    \textcolor{caseblue}{
        \textbf{stable / blocked-for-mapping / unstable-but-visible}
    }
    and adds a per-field
    \textcolor{caseblue}{
        \textbf{fallback-eligibility flag: safe-copy vs.\ block/hide}
    }.
    Reporting uses this additional flag to distinguish fields that can be
    safely displayed from those that should not be rendered or trusted.

    \hfill
    \textcolor{casegreen}{\textbf{Correct}}

    \end{casebox}

    \caption{
    \textbf{Qualitative example of recovering a newly introduced constraint.}
    Global retrieval reaches the correct field-tagging discussion but misses
    \texttt{Msg\_6579}, which changes the policy from a state tag alone to a
    two-part rule containing both display state and fallback eligibility.
    Temporal Evidence Discovery recovers this low-ranked update from the
    corresponding local episode. Blue text highlights the answer-critical
    constraint recovered through temporal locality.
    }

    \label{fig:case_field_constraint}

\end{figure}

\begin{figure}[!t]
    \centering

    \begin{casebox}{
        Temporal Memory Structuring Separates Revision from Execution Constraints
    }

    \footnotesize

    \textbf{Instance.}
    \texttt{Finance / knowledge\_update\_31}

    \vspace{1.5mm}

    \textbf{Question.}

    \emph{What is the current approach for handling edits and ownership
    in the compliance policy draft?}

    \vspace{2mm}
    \hrule
    \vspace{2mm}

    \textbf{Flat evidence presentation}

    \vspace{1mm}

    The retrieved discussion contains messages about control-impact clauses,
    Legal/Ops ownership, and subsequent review requirements. In a flat
    presentation, however, the policy revision and the constraints that
    elaborate the revised policy are not explicitly distinguished.

    \vspace{1mm}

    \textcolor{casepink}{\textbf{BM25 (Top-36) answer}}:
    identifies a general division in which Legal handles wording and sign-off
    while Ops handles implementation-related items, but does not recover the
    clause-specific gate or its owner/backup requirements.

    \hfill
    \textcolor{casepink}{\textbf{Incorrect}}

    \vspace{2.5mm}

    \textbf{Temporal Memory Structuring}

    \vspace{1mm}

    \method organizes the evidence into a path that distinguishes a policy
    update from the subsequent constraints that elaborate the new policy:

    \vspace{1.5mm}

    \begin{evidencebox}

    \textbf{\texttt{Msg\_5666} --- preceding control state}

    \vspace{0.8mm}

    Hold the hard block only on control-impact clauses and allow
    non-control edits to continue while the Legal/Ops ownership split
    is finalized.

    \end{evidencebox}

    \vspace{0.8mm}

    \begin{center}
        \textcolor{caseblue}{\textbf{\textsc{Update} $\Downarrow$}}
    \end{center}

    \vspace{0.3mm}

    \begin{evidencebox}

    \textbf{\texttt{Msg\_7186} --- revised policy}

    \vspace{0.8mm}

    The earlier ``freeze triage first, assign ownership after full
    classification'' procedure no longer applies.

    \vspace{0.8mm}

    \textcolor{caseblue}{
        \textbf{New rule:}
        keep non-control edits moving, hard-block only control-impact
        clauses, and assign an owner and backup at the clause level once
        each item is classified.
    }

    \end{evidencebox}

    \vspace{0.8mm}

    \begin{center}
        \textcolor{caseblue}{\textbf{\textsc{Expand} $\Downarrow$}}
    \end{center}

    \vspace{0.3mm}

    \begin{evidencebox}

    \textbf{\texttt{Msg\_9253} --- execution constraints}

    \vspace{0.8mm}

    Keep \texttt{@User\_10} as owner and \texttt{@User\_13} as backup,
    and require a timestamp, clause-specific change summary, and Legal
    re-read before an affected clause moves forward.

    \end{evidencebox}

    \vspace{2mm}

    \textbf{Gold answer}

    \vspace{0.8mm}

    Instead of freezing the full log, use a clause-specific gate:
    allow non-control edits to continue, hard-block only control-impact
    clauses, and assign an owner and backup for each clause after
    classification.

    \vspace{2mm}

    \textbf{\method answer (excerpt)}

    \vspace{0.8mm}

    The current approach is to
    \textcolor{caseblue}{
        \textbf{freeze and tightly control only risky or touched clauses}
    }
    rather than the full draft. The temporal evidence forest also retains the
    clause-level implementation requirements, including an owner, backup,
    timestamp, change summary, and Legal re-read before affected wording
    proceeds.

    \hfill
    \textcolor{casegreen}{\textbf{Correct}}

    \end{casebox}

    \caption{
    \textbf{Qualitative example combining \textsc{Update} and
    \textsc{Expand} relations.}
    The temporal evidence forest distinguishes the revision of the control policy
    from the later message that adds implementation constraints.
    \textsc{Update} captures the transition to a clause-specific gate,
    whereas \textsc{Expand} attaches ownership and review requirements to
    the revised state. Blue text highlights the current policy and the
    relations used to organize its supporting evidence.
    }

    \label{fig:case_clause_gate}

\end{figure}

\clearpage

This case further illustrates that topical coverage does not imply state
completeness. The retrieved context may contain extensive information about
field categories while omitting a small but decisive constraint added later
in the same discussion. In this instance, the missing information does not
replace the three-state scheme; it augments it with a second decision
dimension. Recovering the local update is therefore necessary to distinguish
the current two-part rule from an incomplete description based only on the
earlier state labels. The failure mode is subtle because an answer that
mentions the three display states can appear highly relevant while still
missing the policy dimension that determines whether a field may be shown.
By recovering \texttt{Msg\_6579}, Temporal Evidence Discovery restores the
interaction between display state and fallback eligibility rather than merely
adding another related message. This makes the case a concrete example of how
low-ranked updates can change the operational meaning of an otherwise
well-covered topic.

\subsection{Separating Policy Revision from Execution Constraints}
\label{app:case_clause_gate}

Figure~\ref{fig:case_clause_gate} illustrates a more complex temporal
evidence structure containing both state progression and supporting
elaboration. The discussion concerns how edits to a compliance-policy draft
should be controlled. The central decision changes from a broad freeze-based
procedure to a clause-specific gate, after which a later message adds
execution constraints concerning ownership, timestamps, change summaries,
and Legal review. Temporal Memory Structuring separates these two functions
through an \textsc{Update} edge followed by an \textsc{Expand} edge.

This case illustrates that temporal organization involves more than selecting
the most recent message. Different later messages can play different roles:
one may revise the governing state, while another may preserve that state and
add operational detail. Treating all later messages as interchangeable
updates would obscure this distinction. The temporal evidence forest instead represents
\texttt{Msg\_5666 $\rightarrow$ Msg\_7186} as an \textsc{Update} and
\texttt{Msg\_7186 $\rightarrow$ Msg\_9253} as an \textsc{Expand}, allowing
the answer model to recover both the revised clause-specific policy and the
constraints required to execute it. The distinction between the two edge
types is important here because \texttt{Msg\_9253} does not supersede the
clause-specific gate established by \texttt{Msg\_7186}; it specifies how that
gate should be executed. Preserving this separation lets the temporal evidence
forest encode both state precedence and supporting operational detail without
collapsing them into a single notion of recency. The resulting path therefore
provides a more faithful account of how the policy evolves and becomes
actionable over time.

\clearpage
\section{Limitations}
\label{app:limitations}

\paragraph{Method limitations.}
First, Temporal Evidence Discovery relies on the initial Top-\(k\) retrieved messages as anchors 
for identifying temporally dense regions and triggering local evidence search. When the initial 
retrieval budget is small, the anchor set may contain too few relevant messages to form a 
reliable cluster, or may fail to cover the relevant episode altogether. In such cases, temporal 
expansion has limited ability to recover the missing evidence. Therefore, although temporal 
locality can improve retrieval beyond the initial ranking, its effectiveness still depends 
partly on the quality and coverage of the coarse anchors.

Second, Temporal Memory Structuring uses two broad edge types, UPDATE and EXPAND, to 
approximately reconstruct the evolution of an event. This abstraction intentionally 
keeps the resulting memory structure simple, but the two relations may conflate semantically 
distinct discourse functions. For example, questioning, clarification, justification, and 
elaboration may all be represented as EXPAND, even though they play different roles in the 
underlying interaction. Similarly, different forms of revision or correction may be grouped 
under UPDATE. Consequently, the resulting forest captures the coarse progression and supporting 
structure of an event rather than a complete, fine-grained discourse representation. 

Finally, both benchmarks provide cleaner metadata and more controlled interaction histories
than many deployed group-chat systems, so generalization to noisy real-world
conversations requires further study.

\paragraph{Analysis limitations.}
First, Exp-2 evaluates evidence recovery against manually audited canonical evidence and  
identified alternatives, but the set of valid alternatives may still be incomplete. Multi-hop 
questions can admit multiple valid reasoning chains, and not every semantically equivalent 
message or paraphrase can be annotated. Moreover, an LLM may answer correctly using a 
semantically similar message, parametric knowledge, or an unannotated paraphrase even when the 
canonical evidence is absent. The alternative-evidence audit reduces this mismatch but cannot 
guarantee exhaustive coverage of valid evidence. Thus, the evidence-recall results in Exp-2 
should be interpreted as a diagnostic estimate of recoverable answer-supporting evidence rather 
than an exhaustive measure of all possible valid evidence.

Second, the final evaluation on both benchmarks depends on LLM-based answer generation and 
judging. LLM outputs can exhibit stochastic variation, and the judge may also respond 
differently to semantically equivalent, partially correct, or underspecified answers. This 
variability in generation and evaluation can introduce noise into the reported accuracy and may 
cause small variations across repeated runs. Although we use consistent answer requirements, 
judging protocols, and decoding settings within controlled comparisons, these controls cannot 
completely eliminate model-induced evaluation variance.

\clearpage
\onecolumn
\clearpage
\section{Answer and Judge Prompts}
\label{app:answer_judge_prompts}

Figure~\ref{fig:prompt_answer_judge} compares the two answer interfaces. The
flat and forest variants share the same evidence-only answering constraints and
the same judge prompt. Their only material difference is evidence
presentation: the flat prompt supplies messages in retrieval-rank order,
whereas the forest prompt exposes tree boundaries and
\textsc{Root}/\textsc{Update}/\textsc{Expand} labels. Blue text marks these
forest-specific instructions. This controlled change isolates presentation and
organization from judge behavior.

\begin{figure}[!ht]
  \centering
  \begin{casebox}{Answer and Judge Prompts: Flat vs. Temporal Forest Presentation}
  \footnotesize
  \begin{minipage}[t]{0.485\linewidth}
    \textbf{Flat Evidence Presentation}

    \vspace{1.2mm}
    \textbf{System Prompt}

    You answer questions using only the supplied retrieved messages. Treat the
    message content as the sole factual source. Do not invent facts, do not
    expose hidden reasoning, and do not mention the prompt, retrieval, or
    judging process. Return only a direct, concise answer.

    \vspace{1mm}
    \textbf{Shared User Prompt Template}
    \vspace{1mm}
    \begin{evidencebox}
      \scriptsize\ttfamily
      Asking user: \textless asking\_user\_id\textgreater\par
      Question:\par
      \textless question\textgreater\par
      Retrieved conversation passages:\par
      \textless serialized\_retrieved\_messages\textgreater\par
      Answer the question using only the retrieved conversation passages above.
    \end{evidencebox}
  \end{minipage}
  \hfill
  \begin{minipage}[t]{0.485\linewidth}
    \textcolor{caseblue}{\textbf{Temporal Forest Evidence Presentation}}

    \vspace{1.2mm}
    \textbf{System Prompt}

    You answer questions using only the supplied retrieved messages.

    {\color{caseblue}
    The messages are organized into event-evolution trees and serialized in
    preorder:
    \begin{itemize}
      \setlength{\itemsep}{0.3mm}
      \setlength{\parskip}{0pt}
      \item Each \texttt{TREE} represents a separate event progression.
      \item Indentation indicates parent--child depth.
      \item \texttt{ROOT} begins an event progression.
      \item \texttt{UPDATE} marks a change to the parent state.
      \item \texttt{EXPAND} adds support or clarification without replacing the
      parent state.
      \item Use timestamps and these relations to reconstruct the relevant
      progression, following \texttt{UPDATE} chains to identify the current or
      final state.
    \end{itemize}}

    Treat the message content as the sole factual source. Do not invent facts,
    do not expose hidden reasoning, and do not mention the prompt, retrieval, or
    judging process. Return only a direct, concise answer.

    \vspace{1mm}
    \textbf{Shared User Prompt Template}
    \vspace{1mm}
    \begin{evidencebox}
      \scriptsize\ttfamily
      Asking user: \textless asking\_user\_id\textgreater\par
      Question:\par
      \textless question\textgreater\par
      Retrieved conversation passages:\par
      \textless serialized\_retrieved\_messages\textgreater\par
      Answer the question using only the retrieved conversation passages above.
    \end{evidencebox}
  \end{minipage}

  \vspace{1.8mm}
  \hrule
  \vspace{1.8mm}
  \textbf{Shared Judge System Prompt.}
  Judge whether the agent answer is semantically consistent with the gold
  answer, accepting meaning-equivalent paraphrases. Output only
  \texttt{Final: Correct} or \texttt{Final: Incorrect}.

  \textbf{Shared Judge User Prompt Template.}
  Question: \textless question\textgreater{} $\mid$ Gold Answer:
  \textless gold\_answer\textgreater{} $\mid$ Agent Answer:
  \textless agent\_answer\textgreater
  \end{casebox}
  \caption{Answer and judge prompts for flat and temporal-forest evidence.
  Blue text denotes instructions present only in the forest interface; the judge
  prompt is shared.}
  \label{fig:prompt_answer_judge}
\end{figure}

\clearpage
\section{Temporal Evidence Discovery Prompt}
\label{app:retrieval_prompt}

Evidence Utility Selection in Figure~\ref{fig:prompt_retrieval} operates only on
messages preselected from one topic--time window. It accepts decisive evidence
and strictly necessary support, while rejecting messages that are merely
topically or lexically similar, obsolete, speculative, or insufficient. The
gate is not allowed to answer the question or inspect the gold answer.

\begin{figure}[!ht]
  \centering
  \begin{casebox}{Prompt for Temporal Evidence Discovery}
  \footnotesize
  \textit{Strict query-conditioned evidence selection inside a topic--time window.}

  \vspace{1.5mm}
  \textbf{Inputs}
  \begin{enumerate}
    \setlength{\itemsep}{0.3mm}
    \setlength{\parskip}{0pt}
    \item One user question.
    \item Candidate messages drawn from one topic--time window after embedding preselection.
  \end{enumerate}

  \textbf{Select only}
  \begin{enumerate}
    \setlength{\itemsep}{0.3mm}
    \setlength{\parskip}{0pt}
    \item \textcolor{caseblue}{\textbf{KEY:}} Messages that directly state an
    answer-bearing fact or make a decisive contribution to the answer.
    \item \textcolor{caseblue}{\textbf{SUPPORT:}} For current decisions or
    states, prefer the newest applicable adopted change and retain necessary
    scope, condition, exception, owner, deadline, or implementation status.
  \end{enumerate}

  \textbf{Reject}
  \begin{enumerate}
    \setlength{\itemsep}{0.3mm}
    \setlength{\parskip}{0pt}
    \item \textcolor{casepink}{\textbf{IRRELEVANT:}} Topic similarity, shared
    entities, generic background, motivation, examples, and messages that merely
    help locate another message.
    \item \textcolor{casepink}{\textbf{IRRELEVANT:}} Replaced decisions,
    abandoned intermediate states, and unadopted proposals when the question
    asks for the operative state.
    \item Weak or speculative evidence; an empty selection is preferable.
  \end{enumerate}

  \textbf{Hard constraints}
  \begin{enumerate}
    \setlength{\itemsep}{0.3mm}
    \setlength{\parskip}{0pt}
    \item Use only the supplied question and messages; do not access the gold
    answer or outside history.
    \item Do not answer the question and do not return labels for rejected messages.
  \end{enumerate}

  \begin{evidencebox}
    \textbf{Required output}

    \scriptsize\ttfamily
    \{\"selected\": [\{\"message\_id\": \"Msg\_123\",\\
    \hspace*{9mm}\"reason\": \"Exact answer-bearing fact.\"\}]\}
  \end{evidencebox}
  \end{casebox}
  \caption{Prompt for strict, query-conditioned evidence selection within a
  topic--time window.}
  \label{fig:prompt_retrieval}
\end{figure}

\clearpage
\section{Temporal Memory Structuring Prompt}
\label{app:forest_prompt}

Figure~\ref{fig:prompt_forest} specifies direct-parent selection. The query
narrows relevance but cannot create a relation unsupported by the two
messages. The direct-relation gate, deterministic ordering constraints, and
single-parent rule jointly prevent topical overlap from producing an
artificially dense graph.

\begin{figure}[!ht]
  \centering
  \begin{casebox}{Prompt for Temporal Memory Structuring}
  \footnotesize
  \textit{Query-conditioned direct-parent selection with deterministic temporal-order gates.}

  \vspace{1.5mm}
  \textbf{Task}
  \begin{enumerate}
    \setlength{\itemsep}{0.3mm}
    \setlength{\parskip}{0pt}
    \item For each later child, select at most one earlier direct parent or
    \textsc{Root}; evaluate $e_{ij}=r(q,\text{earlier message},\text{later message})$.
    \item The question narrows relevance but cannot manufacture a relation that
    is not explicit between the two messages.
  \end{enumerate}

  \textbf{Absolute direct-relation gate}
  \begin{enumerate}
    \setlength{\itemsep}{0.3mm}
    \setlength{\parskip}{0pt}
    \item Require the same concrete object, direct continuation, no hidden
    bridge, and counterfactual dependence.
    \item Choose \textsc{Root} when all candidates are connected only by topic,
    chronology, shared vocabulary, broad relevance, or an unstated intermediate event.
  \end{enumerate}

  \textbf{Relation labels}
  \begin{enumerate}
    \setlength{\itemsep}{0.3mm}
    \setlength{\parskip}{0pt}
    \item \textsc{Update}: a direct state transition of the same decision, plan,
    owner, deadline, threshold, scope, implementation status, validation result,
    or operative rule.
    \item \textsc{Expand}: a direct answer, confirmation, rejection, explanation,
    evidence, clarification, or implementation detail for the parent's concrete state.
    \item \textsc{Root}/\textsc{None}: no justified direct semantic parent.
  \end{enumerate}

  \textbf{Hard ordering constraints}
  \begin{enumerate}
    \setlength{\itemsep}{0.3mm}
    \setlength{\parskip}{0pt}
    \item Parent timestamp must be earlier; stage and \texttt{phase\_order} must
    be nondecreasing; within the same stage and phase, \texttt{phase\_progress}
    must be nondecreasing.
    \item Cross-phase edges remain eligible only when all ordering gates and the
    direct-continuity test hold.
  \end{enumerate}

  \begin{evidencebox}
    \textbf{Required output}

    \scriptsize\ttfamily
    \{\"choices\": [\{\"child\_id\": \"Msg\_2\", \"parent\_id\": \"Msg\_1\",\\
    \hspace*{9mm}\"relation\": \"UPDATE|EXPAND|NONE\",\\
    \hspace*{9mm}\"confidence\": 0.91, \"query\_relevance\": 0.83\}]\}
  \end{evidencebox}
  \end{casebox}
  \caption{Prompt for query-conditioned Temporal Memory Structuring.}
  \label{fig:prompt_forest}
\end{figure}

\clearpage

\end{document}